\documentclass[10pt,twocolumn,letterpaper]{article}

\usepackage[pagenumbers]{cvpr} 

\usepackage{bm}
\usepackage{multirow}

\definecolor{cvprblue}{rgb}{0.21,0.49,0.74}
\usepackage[pagebackref,breaklinks,colorlinks,allcolors=cvprblue]{hyperref}

\def\paperID{14438} 
\def\confName{CVPR}
\def\confYear{2025}

\title{PaRCE: Probabilistic and Reconstruction-based Competency Estimation for CNN-based Image Classification}

\author{Sara Pohland\\
University of California, Berkeley\\
Berkeley, CA, USA 94720\\
{\tt\small spohland@berkeley.edu}
\and
Claire Tomlin\\
University of California, Berkeley\\
Berkeley, CA, USA 94720\\
{\tt\small tomlin@berkeley.edu}
}

\begin{document}
\maketitle
\begin{abstract}

Convolutional neural networks (CNNs) are extremely popular and effective for image classification tasks but tend to be overly confident in their predictions. Various works have sought to quantify uncertainty associated with these models, detect out-of-distribution (OOD) inputs, or identify anomalous regions in an image, but limited work has sought to develop a holistic approach that can accurately estimate perception model confidence across various sources of uncertainty. We develop a probabilistic and reconstruction-based competency estimation (PaRCE) method and compare it to existing approaches for uncertainty quantification and OOD detection. We find that our method can best distinguish between correctly classified, misclassified, and OOD samples with anomalous regions, as well as between samples with visual image modifications resulting in high, medium, and low prediction accuracy. We describe how to extend our approach for anomaly localization tasks and demonstrate the ability of our approach to distinguish between regions in an image that are familiar to the perception model from those that are unfamiliar. We find that our method generates interpretable scores that most reliably capture a holistic notion of perception model confidence.

\end{abstract}    
\section{Introduction}
\label{sec:intro}

Convolutional neural networks (CNNs) are very useful in image classification tasks but can fail dramatically with little explanation or warning \cite{saleem_explaining_2022,guo,ovadia,Nguyen_2015_CVPR}. To safely deploy CNN-based classification models in real-world systems, it is crucial to accurately assess the confidence level of their predictions. Extensive research has aimed to quantify uncertainty in CNN predictions \cite{gawlikowski_survey_2023}, but these methods are often overly confident for out-of-distribution (OOD) inputs that differ significantly from those seen during training \cite{schwaiger_is_nodate}. This has led to work focused on detecting inputs outside of the model's training distribution \cite{yang_generalized_2022} and identifying anomalous regions in an image \cite{tao_deep_2022,haber_anomaly_2022,anoopa_survey_2022}. While these methods help identify unfamiliar data, they rely on arbitrary thresholds to make binary decisions about whether an input is anomalous and lack nuanced probabilistic insights to capture the full scope of predictive uncertainty.



\begin{figure*}[h!]
  \centering
  \includegraphics[width=0.83\linewidth]{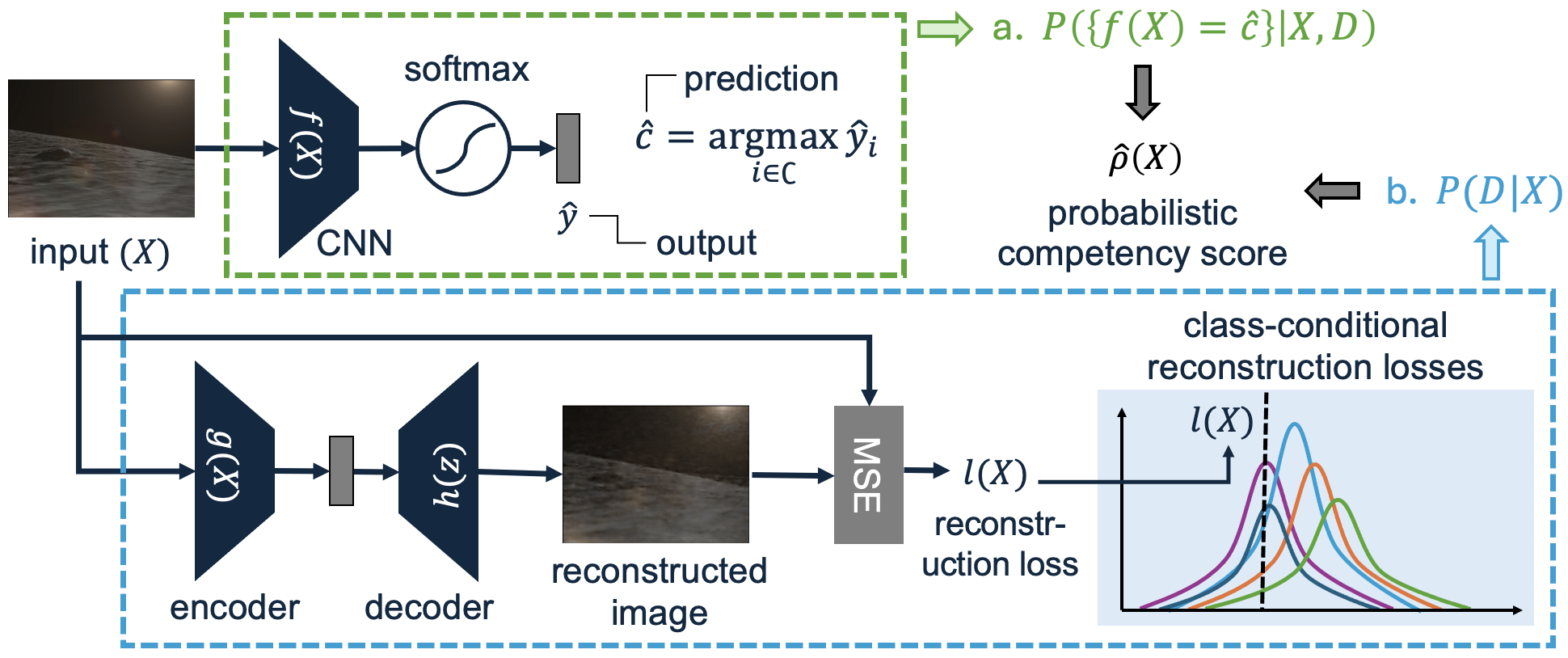}
  \caption{An overview of our PaRCE method. We estimate (a) the probability that the convolutional neural network (CNN) prediction is correct given that the input is in-distribution (ID) based on the classifier output and (b) the probability that the input is ID given the reconstruction loss of an autoencoder trained in parallel with the classifier.}
  \label{fig:overview}
\end{figure*}

We propose a probabilistic and reconstruction-based competency estimation (PaRCE) method. Our method integrates multiple aspects of predictive uncertainty into a single, comprehensive score that reflects the probability that a perception model’s prediction is accurate for a given image. The probabilistic nature of this score increases its interpretability and allows it to be more easily incorporated into a probabilistic decision-making framework that may consider several sources of uncertainty. Additionally, we extend our method to estimate a model’s competency at the regional level within an image, providing valuable information for human users and decision-making systems. To evaluate PaRCE, we introduce several new datasets and evaluation schemes that are motivated by the autonomous navigation of robots in complex, outdoor environments. We compare the performance of our method to existing uncertainty quantification, OOD detection, and anomaly localization methods. Our results demonstrate that PaRCE provides more reliable and interpretable competency estimates across various image types, giving it the potential to enhance model trustworthiness and practical applicability.
\section{Background \& Related Work}
\label{sec:background}

CNNs for image classification usually output softmax scores, which can be interpreted as the probability that an image belongs to each of the training classes. The maximum softmax probability can serve as a measure of model confidence, but these probabilities tend to be very close to one \cite{guo} and are particularly unreliable for OOD data \cite{ovadia}. These scores can be better-calibrated through temperature scaling \cite{guo} and other methods, but these calibrated scores still tend to be overly confident. This has motivated many other approaches to quantify model uncertainty (\cref{subsec:uq}). These methods focus on data/aleatoric uncertainty, arising from complexities of the data (i.e., noise, class overlap, etc.), and model/epistemic uncertainty, reflecting the ability of the CNN to capture the true underlying model of the data \cite{yarin_gal_uncertainty_2016}. These methods do not tend to capture distributional uncertainty resulting from mismatched training and test distributions \cite{quinonero-candela_dataset_2009}. This has led to methods that seek to detect inputs that are OOD (\cref{subsec:ood}). There is also related work that aims to identify and localize anomalous regions in input images that differ from training images (\cref{subsec:anomaly}).

\subsection{Uncertainty Quantification (UQ)}
\label{subsec:uq}

Extensive research has focused on understanding and quantifying uncertainty in a neural network's prediction. The modeling of these uncertainties can be divided into methods based on (1)~Bayesian neural networks (BNNs), (2)~deterministic  networks, and (3)~ensembles of  networks \cite{gawlikowski_survey_2023}. A BNN is a stochastic model whose output is a probability distribution over its predictions \cite{neal-1992,neal_bayesian_1996}. Approaches that employ BNNs extract uncertainty as a statistical measure over the outputs of the model. Methods that use a single deterministic network often rely on Monte Carlo (MC) dropout as approximate Bayesian inference \cite{dropout}. Ensemble methods combine the predictions of multiple deterministic networks to form a probability density function \cite{lakshminarayanan_simple_2017}. While these methods address the typical overconfidence in neural network predictions with better-calibrated confidence scores, they generally focus on predictions for in-distribution inputs, which come from the same distribution as the training data. These approaches are generally insufficient to appropriately assign confidence scores for OOD inputs that differ significantly from those seen during training \cite{schwaiger_is_nodate}.

\subsection{Out-of-Distribution (OOD) Detection}
\label{subsec:ood}

Many recent approaches have focused on quantifying distributional uncertainty and determining if an input falls outside of the input-data distribution. These approaches are generally either (1)~classification-based, (2)~density-based, (3)~distance-based, or (4)~reconstruction-based \cite{yang_generalized_2022}. Classification-based methods seek to revise the overly confident softmax scores at the output of neural networks to detect OOD samples more robustly \cite{liu_energy-based_2020,liang_enhancing_2020,openmax,dice}. Density-based methods model the training distribution with some probabilistic model and flag test data in low-density regions as OOD \cite{zong2018deep,rezende_variational_2016,kingma_glow_2018,ren-2019}. Distance-based methods use distance metrics in the feature space with the assumption that OOD samples should lie relatively far from the centroids or prototypes of the training classes \cite{kl_matching,lee-2018,sun_out--distribution_2022}. Finally, reconstruction-based methods rely on the reconstruction loss of autoencoders (AEs) or generative adversarial networks (GANs), assuming OOD samples will be reconstructed poorly \cite{xia-2015,gong_memorizing_2019,Sabokrou}. While all these methods have successfully identified OOD samples, they do not fully capture the predictive uncertainty associated with CNNs.

\subsection{Anomaly Detection \& Localization}
\label{subsec:anomaly}

A field of study closely related to OOD detection is anomaly detection and localization, wherein the goal is to segment the particular pixels containing anomalies \cite{tao_deep_2022,haber_anomaly_2022,narayanan_survey_2023,anoopa_survey_2022}. Within the area of anomaly localization, most approaches are (1)~reconstruction-based, (2)~synthesizing-based, or (3)~embedding-based \cite{liu_simplenet_2023}. Reconstruction-based methods use pixel-wise reconstruction errors to detect anomalies under the assumption that networks trained with only ``normal" data will not accurately reconstruct anomalous regions \cite{bergmann_improving_2019,zavrtanik_reconstruction_2021,ganomaly}. Synthesizing-based methods estimate the decision boundary between normal and anomalous data by training on synthetic anomalies generated from anomaly-free images \cite{zavrtanik_draem_2021,li_cutpaste_2021}. Embedding-based methods embed image features into a compressed space and assume that anomalous features are far from the normal clusters in the embedding space \cite{del_bimbo_padim_2021,patchcore,reverse,stfpm,fastflow}. Similar to the OOD detection methods, these methods can successfully identify anomalies in an image but cannot fully capture uncertainty.

\subsection{Probabilistic Competency Estimation}

UQ techniques focus on quantifying data/aleatoric uncertainty and model/epistemic uncertainty but generally fail to capture uncertainty arising from distributional shift. OOD and anomaly detection techniques better address this type of uncertainty but generally rely on thresholds to generate a binary decision, rather than capturing a holistic measure of uncertainty. We are interested in \textit{perception model competency}--a generalized form of predictive uncertainty introduced by Rajendran and LeVine \cite{alice}. In this work, a competency score is an estimated lower bound on the probability that the classifier error is below some threshold. We develop a new method for estimating model competency.

\section{Definition of PaRCE Score}
\label{sec:approach}

We define a probabilistic score that captures model, data, and distributional uncertainty arising in a CNN-based perception model (\cref{subsec:competency}), using a reconstruction-based approach to estimate distributional uncertainty (\cref{subsec:id-prob}). Our method is designed to directly reflect the prediction accuracy of the perception model (\cref{subsec:calibration}) and can be extended to generate regional competency images (\cref{subsec:images}).

\subsection{Estimating Model Competency}
\label{subsec:competency}

Let $f$ be the true underlying model of the system from which our images are drawn and $\hat{f}$ be the predicted model (referred to as the perception model or classifier). For an input image, $\bm{X}$, the perception model aims to estimate the true class of the image, $f(\bm{X})$, from the set of all classes, $\mathcal{C}$. The competency of the model for this image is given by
\begin{equation}
    \rho(\bm{X}) := P\bigl(\{\hat{f}(\bm{X})=f(\bm{X})\}|\bm{X}\bigr).
    \label{eq:competency}
\end{equation}
To simplify our notation, let $\hat{c}$ be the class predicted by the perception model (i.e., $\hat{f}(\bm{X})=\hat{c}$) such that
\begin{equation}
    \rho(\bm{X}) = P\bigl(\{f(\bm{X})=\hat{c}\}|\bm{X}\bigr).
    \label{eq:competency-simplified}
\end{equation}

Often, the perception model uses the softmax function to obtain an estimate of the probability $P(\{f(\bm{X})=c)\}|\bm{X})$ for each class $c\in\mathcal{C}$. However, the perception model cannot truly estimate this probability because it is limited by the data contained in the training sample. It instead estimates the probability $P(\{f(\bm{X})=c)\}|\bm{X}, D)$, where $D$ is the event that the input image is in-distribution (ID) (i.e., drawn from the same distribution as the training samples). Let us then write the following lower bound on competency:
\begin{align}
    \rho(\bm{X}) 
    &\geq 
    P\bigl(D \cap \{f(\bm{X})=\hat{c})\}|\bm{X}\bigr) \\
    &= P\bigl(\{f(\bm{X})=\hat{c}\}|\bm{X}, D\bigr)P(D|\bm{X}).
\end{align}

As summarized in Figure \ref{fig:overview}, we assume that the perception model provides an estimate of the class probability, but we need a method to estimate the in-distribution probability.

\subsection{Estimating In-Distribution (ID) Probability}
\label{subsec:id-prob}

To estimate the ID probability, we design an autoencoder (AE) to reconstruct the same images used to train the classifier. The AE’s encoder mirrors the classifier’s architecture, while its decoder is the inverse of the classifier. This architecture establishes a direct connection between the classifier’s feature vector and the AE’s latent representation, making reconstruction loss a proxy for the classifier’s familiarity with an input image and the ability of its learned features to capture the input. When the classifier’s features effectively represent an image, the reconstruction loss remains low, indicating higher prediction confidence. Conversely, a high reconstruction loss suggests a discrepancy between the input image and ID data, implying a decrease in confidence.

The probability an image, $\bm{X}$, is drawn from the same distribution as those in the training sample is given by
\begin{equation}
    P(D|\bm{X}) = \sum_{c\in\mathcal{C}} P(D|\{f(\bm{X})=c\})P(\{f(\bm{X})=c\}|\bm{X}).
    \label{eq:id-prob}
\end{equation}
Assume now that the classifier provides an estimate of $P(\{f(\bm{X})=c\}|\bm{X})$. We can estimate $P(D|\{f(\bm{X})=c\})$ as the probability that the reconstruction loss, $\ell(\bm{X})$, for image $\bm{X}$ falls within the lowest $N\%$ of losses from the training images. Using a holdout set, $\mathcal{X}$, randomly selected from the same distribution as the training set but not used to train the classifier, we estimate the distribution of reconstruction loss, $\mathcal{L}_c$, for each class $c\in\mathcal{C}$ as a Gaussian with mean $\mu_c$ and standard deviation $\sigma_c$. Because $\mathcal{L}_c$ follows a Gaussian distribution, $N$ corresponds to a z-score, $z$. We estimate the class in-distribution probability, $P(D|\{f(\bm{X})=c\})$, as 
\begin{equation}
    \hat{P}(D|\{f(\bm{X})=c\}) = 1 - \phi\left(\frac{\ell(\bm{X}) - 2\mu_c}{\sigma_c} - z\right),
    \label{eq:id-class}
\end{equation}
where $\phi$ denotes the cumulative distribution function (CDF) of the standard normal distribution. (See \cref{eq:deriv-1,eq:deriv-2,eq:deriv-3,eq:deriv-4,eq:deriv-5} for the derivation of this estimate.) Let $\hat{p}_c$ be the probabilistic output of the classifier for class $c\in\mathcal{C}$ and $\hat{p}_{\hat{c}}$ be the maximum softmax probability corresponding to the predicted class. We now have the following estimate of competency:
\begin{equation}
    \hat{\rho}(\bm{X}) := \hat{p}_{\hat{c}} \sum_{c\in\mathcal{C}} \hat{p}_c\left(1 - \phi\left(\frac{\ell(\bm{X}) - 2\mu_c}{\sigma_c} - z\right)\right).
    \label{eq:estimated-competency}
\end{equation}

\subsection{Calibrating Competency Score}
\label{subsec:calibration}

In \cref{eq:estimated-competency}, $\hat{p}_c$ is provided by the trained perception model, while $\mu_c$ and $\sigma_c$ are determined by the holdout set drawn from the training distribution. We select the z-score, $z$, such that the average competency score reflects the prediction accuracy for the ID holdout set, $\mathcal{X}$. To calibrate our competency estimator, we want to find $z$ such that 
\begin{equation}
    \frac{1}{|\mathcal{X}|}\sum_{\bm{X}\in\mathcal{X}}\hat{\rho}(\bm{X}) = \frac{1}{|\mathcal{X}|}\sum_{\bm{X}\in\mathcal{X}} 1_{\{\hat{f}(\bm{X})=f(\bm{X})\}}(\bm{X}).
    \label{eq:calibration}
\end{equation}

Rather than computing the exact value of $z$ that satisfies this equation, we select a z-score to the nearest five hundredth that results in the average competency score (left side of \cref{eq:calibration}) closest to the prediction accuracy (right side of \cref{eq:calibration}). (See \cref{fig:zscore} for a plot of the calibration curves.)

\subsection{Generating Competency Images}
\label{subsec:images}

Suppose that, in addition to estimating the overall competency score for the image as a whole, we wish to estimate the competency of a model for particular regions in the image. Now, instead of letting $\bm{X}$ be the entire input image, $\bm{X}$ is a segmented region of the image. In our work, images are segmented using the graph-based image segmentation algorithm  developed by Felzenszwalb \cite{felzenszwalb_efficient_2004}. We follow roughly the same procedures to estimate the probability that each region in the input image came from the same distribution as the training samples. Now, rather than designing an AE to reconstruct the input image, we design an image inpainting model to reconstruct a missing segment of the image and measure the average reconstruction loss over the pixels corresponding to that image segment. We calibrate these scores again to reflect the prediction accuracy of the perception model. (See \cref{fig:zscore-seg} for the calibration curves.) We can then use these scores to generate regional competency images. (See the last column of \cref{fig:regional-lunar,fig:regional-speed,fig:regional-park}.)

\section{Analysis of PaRCE Score}
\label{sec:results}

We first compare our overall PaRCE score (\cref{subsec:competency}) to existing methods for uncertainty quantification (\cref{subsec:uq}) and OOD detection (\cref{subsec:ood}) in \cref{subsec:results-overall}. We then compare our regional PaRCE image (\cref{subsec:images}) to existing methods for anomaly detection and localization (\cref{subsec:anomaly}) in \cref{subsec:results-regional}.

\subsection{Overall Competency Scores}
\label{subsec:results-overall}

We compare our overall model competency estimation method against various existing methods for quantifying uncertainty and detecting OOD inputs. In particular, we consider the Maximum Softmax Probability (MSP) baseline, the calibrated MSP with Temperature Scaling \cite{guo}, Monte Carlo (MC) Dropout \cite{dropout}, Ensembling \cite{lakshminarayanan_simple_2017}, the Energy Score \cite{liu_energy-based_2020}, ODIN \cite{liang_enhancing_2020}, OpenMax \cite{openmax}, DICE \cite{dice}, KL-Matching \cite{kl_matching}, the Mahalanobis Distance \cite{lee-2018}, and k-Nearest Neighbors (k-NN) \cite{sun_out--distribution_2022}. These methods serve as a representative selection of the many existing approaches and are often used as baselines in other works. Note that we focus on the most popular methods that are easily implemented with the help of the PyTorch-OOD library \cite{kirchheim2022pytorch}. 

\subsubsection{Evaluation 1: Manually Identified OOD Samples} 
\label{subsubsec:results-overall-1}

\textbf{Datasets:} We conduct analysis across three unique datasets that contain manually identified OOD samples. The first dataset was obtained from a lunar environment, in which the training set contains images from an uninhabited moon and the test set contains images from a habited moon. The second dataset contains speed limit signs in Germany \cite{gtsdb}. While the training dataset is composed of common speed limit signs (30 km/hr and higher), the test set also contains an uncommon speed limit (20 km/hr). The third dataset contains images from regions in a park, in which the training data only contains images from forested and grassy regions of the park, while the test set additionally includes images from around the park's pavilion. See \cref{sec:app-datasets} for additional details and example images.

\textbf{Metrics:} We evaluate scoring methods based on their computation time and their ability to distinguish between correctly classified, misclassified, and OOD samples. To quantify the ability to distinguish between sets of samples, we consider the distance between score distributions using the Kolmogorov–Smirnov (KS) test, the extent of overlap between distributions as measured by the area under the receiver operating characteristic (AUROC), and the detection error determined by the false positive rate (FPR) at a 95\% true positive rate (TPR), where a true positive indicates the correct identification of a misclassified or OOD sample. See \cref{sec:app-analysis} for additional details on these metrics.

\begin{table*}[h!]
    \centering
    \renewcommand{\arraystretch}{1.2} 
    \resizebox{0.9\textwidth}{!}{%
    \begin{tabular}{|c|r|rrr|rrr|rrr|}
    \hline
    \multirow{2}{*}{Method} & \multirow{2}{*}{\shortstack{Computation \\ Time (sec) $\downarrow$}} & \multicolumn{3}{c|}{Correct vs. Incorrect} & \multicolumn{3}{c|}{Correct vs. OOD} & \multicolumn{3}{c|}{Incorrect vs. OOD} \\
    \cline{3-5} \cline{6-8} \cline{9-11}
     & & Dist. $\uparrow$ & AUROC $\uparrow$ & FPR $\downarrow$ & Dist. $\uparrow$ & AUROC $\uparrow$ & FPR $\downarrow$ & Dist. $\uparrow$ & AUROC $\uparrow$ & FPR $\downarrow$ \\
    \hline
    \renewcommand{\arraystretch}{1} 
    Softmax & \textbf{0.0001} &  0.72 & 0.91 & 1.00 & 0.45 & 0.76 &1.00 & 0.13 & 0.76 & 1.00 \\
    Temperature \cite{guo} & 0.0002 & 0.72 & 0.91 & 1.00 & 0.45 &  0.76 & 1.00 & 0.13 & 0.75 & 1.00 \\
    MC Dropout \cite{dropout} & 0.2700 & 0.62 & 0.86 & 0.47 & 0.46 & 0.80 & 0.72 & 0.05 & 0.79 & 0.73 \\
    Ensemble \cite{lakshminarayanan_simple_2017} & 0.0668 & \textbf{0.80} &  \textbf{0.93} & \textbf{0.28} & 0.49 & 0.79 & 1.00 & 0.01 & 0.78 & 1.00 \\
    Energy \cite{liu_energy-based_2020} & 0.0002 & 0.71 & 0.90 & 1.00 & 0.45 & 0.75 & 1.00 & 0.19 & 0.75 & 1.00 \\
    ODIN \cite{liang_enhancing_2020} & 0.1056 & 0.02 & 0.40 & 1.00 & 0.03 & 0.48 & 1.00 & 0.19 & 0.49 & 1.00 \\
    OpenMax \cite{openmax} & 0.0017 & 0.31 & 0.62 & 0.87 & 0.11 & 0.50 & 0.87 & 0.06 & 0.49 & 0.87 \\
    DICE \cite{dice} & 0.0129 & 0.61 & 0.87 & 0.50 & 0.31 & 0.72 & 0.81 & 0.11 & 0.71 & 0.81 \\
    KL-Matching \cite{kl_matching} & 0.0004 & 0.65 & 0.80 & 0.83 & 0.44 & 0.74 & 0.88 & 0.22 & 0.74 & 0.88 \\
    Mahalanobis \cite{lee-2018} & 0.1092 & 0.53 & 0.80 & 0.58 & 0.45 & 0.81 & 0.58 & 0.20 & 0.80 & 0.58 \\
    k-NN \cite{sun_out--distribution_2022} & 0.0190 & 0.59 & 0.83 & 0.41 & 0.61 & 0.87 & 0.37 & 0.29 & 0.87 & 0.38 \\
    \hline
    PaRCE (Ours) & 0.0270 & 0.64 & 0.87 & 0.59 & \textbf{0.89} & \textbf{0.99} & \textbf{0.08} & \textbf{0.71} & \textbf{0.99} & \textbf{0.09} \\
    \hline
    \end{tabular}}
    \caption{Comparison of measures of overall model competency for the lunar dataset with manually identified OOD samples.}
    \label{tab:overall-lunar-1}
\end{table*}

\begin{table*}[h!]
    \centering
    \renewcommand{\arraystretch}{1.2} 
    \resizebox{0.9\textwidth}{!}{%
    \begin{tabular}{|c|r|rrr|rrr|rrr|}
    \hline
    \multirow{2}{*}{Method} & \multirow{2}{*}{\shortstack{Computation \\ Time (sec) $\downarrow$}} & \multicolumn{3}{c|}{Correct vs. Incorrect} & \multicolumn{3}{c|}{Correct vs. OOD} & \multicolumn{3}{c|}{Incorrect vs. OOD} \\
    \cline{3-5} \cline{6-8} \cline{9-11}
     & & Dist. $\uparrow$ & AUROC $\uparrow$ & FPR $\downarrow$ & Dist. $\uparrow$ & AUROC $\uparrow$ & FPR $\downarrow$ & Dist. $\uparrow$ & AUROC $\uparrow$ & FPR $\downarrow$ \\
    \hline
    \renewcommand{\arraystretch}{1} 
    Softmax & \textbf{0.0001} & \textbf{0.86} & \textbf{0.97} & \textbf{0.10} & \textbf{0.78} & \textbf{0.94} & 0.18 & 0.06 & \textbf{0.93} & 0.20 \\
    Temperature \cite{guo} & 0.0002 & \textbf{0.86} & \textbf{0.97} & \textbf{0.10} & \textbf{0.78} & \textbf{0.94} & 0.18 & 0.06 & \textbf{0.93} & 0.20 \\
    MC Dropout \cite{dropout} & 6.7872 & 0.83 & 0.94 & 0.17 & \textbf{0.78} & 0.93 & \textbf{0.17} & 0.15 & \textbf{0.93} & 0.20 \\
    Ensemble \cite{lakshminarayanan_simple_2017} & 1.5363 & 0.77 & 0.90 & 0.19 & 0.68 & 0.90 & 0.28 & 0.17 & 0.89 & 0.31 \\
    Energy \cite{liu_energy-based_2020} & 0.0002 & \textbf{0.86} & \textbf{0.97} & \textbf{0.10} & \textbf{0.78} & \textbf{0.94} & 0.18 & 0.03 & \textbf{0.93} & 0.20 \\
    ODIN \cite{liang_enhancing_2020} & 0.8650 & 0.41 & 0.70 & 1.00 & 0.20 & 0.55 & 1.00 & 0.00 & 0.53 & 1.00 \\
    OpenMax \cite{openmax} & 0.0012 & 0.53 & 0.77 & 0.83 & 0.14 & 0.48 & 0.84 & 0.03 & 0.45 & 0.84 \\
    DICE \cite{dice} & 0.3228 & 0.74 & 0.92 & 0.22 & 0.77 & 0.92 & 0.21 & 0.16 & 0.91 & 0.21 \\
    KL-Matching \cite{kl_matching} & 0.0004 & \textbf{0.86} & 0.96 & \textbf{0.10} & 0.76 & \textbf{0.94} & 0.18 & 0.00 & \textbf{0.93} & 0.20 \\
    Mahalanobis \cite{lee-2018} & 0.8937 & 0.34 & 0.66 & 0.72 & 0.50 & 0.78 & 0.67 & \textbf{0.38} & 0.79 & 0.66 \\
    k-NN \cite{sun_out--distribution_2022} & 0.3330 & 0.45 & 0.74 & 0.60 & 0.55 & 0.83 & 0.45 & 0.28 & 0.84 & 0.44 \\
    \hline
    PaRCE (Ours) & 0.0928 & 0.85 & \textbf{0.97} & 0.11 & \textbf{0.78} & \textbf{0.94} & 0.18 & 0.07 & \textbf{0.93} & \textbf{0.19} \\
    \hline
    \end{tabular}}
    \caption{Comparison of measures of overall model competency for the speed dataset with manually identified OOD samples.}
    \label{tab:overall-speed-1}
\end{table*}

\begin{table*}[h!]
    \centering
    \renewcommand{\arraystretch}{1.2} 
    \resizebox{0.9\textwidth}{!}{%
    \begin{tabular}{|c|r|rrr|rrr|rrr|}
    \hline
    \multirow{2}{*}{Method} & \multirow{2}{*}{\shortstack{Computation \\ Time (sec) $\downarrow$}} & \multicolumn{3}{c|}{Correct vs. Incorrect} & \multicolumn{3}{c|}{Correct vs. OOD} & \multicolumn{3}{c|}{Incorrect vs. OOD} \\
    \cline{3-5} \cline{6-8} \cline{9-11}
     & & Dist. $\uparrow$ & AUROC $\uparrow$ & FPR $\downarrow$ & Dist. $\uparrow$ & AUROC $\uparrow$ & FPR $\downarrow$ & Dist. $\uparrow$ & AUROC $\uparrow$ & FPR $\downarrow$ \\
    \hline
    \renewcommand{\arraystretch}{1} 
    Softmax & \textbf{0.0001} & 0.35 & 0.67 & 1.00 & 0.33 & 0.69 & 1.00 & 0.07 & 0.69 & 1.00 \\
    Temperature \cite{guo} & 0.0002 & \textbf{0.69} & \textbf{0.87} & \textbf{0.31} & 0.33 & 0.56 & 1.00 & 0.08 & 0.55 & 1.00 \\
    MC Dropout \cite{dropout} & 0.2030 & 0.01 & 0.41 & 0.99 & 0.38 & 0.62 & 0.97 & 0.47 & 0.67 & 0.96 \\
    Ensemble \cite{lakshminarayanan_simple_2017} & 0.0471 & 0.44 & 0.64 & 1.00 & 0.61 & 0.85 & 1.00 & 0.36 & 0.85 & 1.00 \\
    Energy \cite{liu_energy-based_2020} & 0.0002 & 0.35 & 0.65 & 1.00 & 0.31 & 0.67 & 1.00 & 0.04 & 0.67 & 1.00 \\
    ODIN \cite{liang_enhancing_2020} & 0.0621 & 0.48 & 0.72 & 1.00 & 0.00 & 0.46 & 1.00 & 0.02 & 0.46 & 1.00 \\
    OpenMax \cite{openmax} & 0.0018 & 0.17 & 0.44 & 1.00 & 0.54 & 0.75 & 0.44 & 0.80 & 0.76 & 0.44 \\
    DICE \cite{dice} & 0.0109 & 0.26 & 0.49 & 0.80 & 0.07 & 0.34 & 0.90 & 0.03 & 0.34 & 0.90 \\
    KL-Matching \cite{kl_matching} & 0.0004 & 0.24 & 0.32 & 1.00 & 0.42 & 0.72 & 0.57 & 0.72 & 0.73 & 0.57 \\
    Mahalanobis \cite{lee-2018} & 0.0727 & 0.49 & 0.71 & 0.75 & 0.84 & 0.96 & 0.27 & 0.88 & 0.96 & 0.26 \\
    k-NN \cite{sun_out--distribution_2022} & 0.0141 & 0.31 & 0.54 & 1.00 & 0.89 & 0.97 & 0.10 & 0.79 & 0.98 & 0.09 \\
    \hline
    PaRCE (Ours) & 0.0234 & 0.34 & 0.58 & 0.73 & \textbf{0.94} & \textbf{0.98} & \textbf{0.03} & \textbf{0.91} & \textbf{0.99} & \textbf{0.04} \\
    \hline
    \end{tabular}}
    \caption{Comparison of measures of overall model competency for the park dataset with manually identified OOD samples.}
    \label{tab:overall-park-1}
\end{table*}

\textbf{Results:} The results are summarized in \cref{tab:overall-lunar-1,tab:overall-speed-1,tab:overall-park-1}. We display the distribution of competency scores for correctly classified, misclassified, and OOD samples in \cref{fig:box-eval1-lunar,fig:box-eval1-speed,fig:box-eval1-park} in Appendix \ref{sec:app-analysis}. 

The MSP baseline is the simplest and fastest method, but it tends to assign scores of one or close to one regardless of whether an image is correctly classified, misclassified, or OOD, implying nearly complete confidence across sample types. Temperature Scaling is another fast and relatively simple method that seeks to calibrate the softmax scores, but the distributions of scores for this method tend to look quite similar to those of the original softmax method, resulting in an insignificant improvement overall. These methods are sufficient to distinguish correctly classified samples from misclassified and OOD samples in the speed dataset, but they do not perform as well for other evaluations.

MC Dropout and Ensembling are related methods commonly used for uncertainty quantification. The performance of these methods is inconsistent across datasets. Both methods generally assign lower scores to misclassified and OOD samples, as compared to correctly classified ones, but this is not always the case. These methods are generally unable to distinguish between misclassified and OOD samples and often assign high scores to OOD images. Note that Dropout is by far the slowest method we consider, and Ensembling is also on the slower side. Furthermore, Ensembling requires significant computational effort during training.

The classification-based OOD detection techniques--Energy, ODIN, OpenMax, and DICE--have varying performance across datasets, but they generally never outperform other methods we consider. The Energy Score tends to be very similar to the Softmax and Temperature Scaling scores. ODIN assigns a maximum score to nearly all points across all datasets. OpenMax and DICE show variations in score distributions for correctly classified, misclassified, and OOD sets but produce scores with a lot of distribution overlap and little difference in average scores. 

The distance-based OOD detection techniques--KL-Matching, Mahalanobis, and k-NN--tend to assign lower scores to misclassified and OOD images than to correctly classified images, but the extent of this difference varies across each method and dataset. KL-matching does a good job distinguishing correctly classified images from misclassified and OOD ones, but these three methods generally do not outperform other methods we consider.

Our method, PaRCE, performs the most reliably, outperforming other methods across the majority of evaluations. It is generally not the best at distinguishing between correctly classified and misclassified samples but tends to best distinguish correctly classified and misclassified samples from OOD samples. Note that while all methods, other than the MSP, generate scores with somewhat arbitrary values, our method is probabilistic, generating intuitive scores between zero and one. Our method tends to assign scores very close to one to correctly classified samples, lower scores to misclassified samples, and even lower scores to OOD samples. 


\subsubsection{Evaluation 2: Generated Data with New Properties}
\label{subsubsec:results-overall-2}

\textbf{Datasets:} In addition to the spatial anomalies considered in the previous evaluation, we are interested in non-regional anomalies that reduce model confidence and prediction accuracy. For each dataset discussed in \cref{subsubsec:results-overall-1}, we generate new data from the ID samples by adjusting saturation, contrast, brightness, pixelation, or noise levels. We consider 21 values for each of these image properties, each of which results in either high, medium, or low prediction accuracy. See \cref{sec:app-datasets} for additional details and examples.

\textbf{Metrics:} We evaluate methods based on their ability to distinguish between image modifications resulting in high, medium, and low accuracy. To evaluate the ability to distinguish between sets of samples using each scoring method, we consider the same metrics discussed in \cref{subsubsec:results-overall-1}.

\begin{table*}[h!]
    \centering
    \renewcommand{\arraystretch}{1.2} 
    \resizebox{0.8\textwidth}{!}{%
    \begin{tabular}{|c|rrr|rrr|rrr|}
    \hline
    \multirow{2}{*}{Method}  & \multicolumn{3}{c|}{High vs. Medium} & \multicolumn{3}{c|}{High vs. Low} & \multicolumn{3}{c|}{Medium vs. Low} \\
    \cline{2-4} \cline{5-7} \cline{8-10}
     & Dist. $\uparrow$ & AUROC $\uparrow$ & FPR $\downarrow$ & Dist. $\uparrow$ & AUROC $\uparrow$ & FPR $\downarrow$ & Dist. $\uparrow$ & AUROC $\uparrow$ & FPR $\downarrow$ \\
    \hline
    \renewcommand{\arraystretch}{1} 
    Softmax & 0.16 & 0.59 & 1.00 & 0.08 & 0.57 & 1.00 & 0.00 & 0.51 & 1.00 \\
    Temperature \cite{guo} & 0.24 & 0.63 & \textbf{0.75} & 0.24 & 0.62 & 0.85 & 0.07 & 0.56 & 0.90 \\
    MC Dropout \cite{dropout} & 0.23 & 0.64 & 1.00 & 0.30 & 0.63 & 1.00 & 0.08 & 0.57 & 1.00 \\
    Ensemble \cite{lakshminarayanan_simple_2017} & 0.31 & 0.69 & 1.00 & 0.36 & 0.68 & 1.00 & 0.12 & 0.60 & 1.00 \\
    Energy \cite{liu_energy-based_2020} & 0.15 & 0.59 & 1.00 & 0.08 & 0.57 & 1.00 & 0.00 & 0.52 & 1.00 \\
    ODIN \cite{liang_enhancing_2020} & 0.00 & 0.46 & 1.00 & 0.00 & 0.43 & 1.00 & 0.00 & 0.43 & 1.00 \\
    OpenMax \cite{openmax} & 0.12 & 0.55 & 0.92 & 0.26 & 0.58 & 0.92 & 0.15 & 0.56 & 0.93 \\
    DICE \cite{dice} & 0.08 & 0.53 & 0.98 & 0.02 & 0.46 & 1.00 & 0.00 & 0.41 & 1.00 \\
    KL-Matching \cite{kl_matching} & 0.19 & 0.59 & 0.96 & 0.46 & 0.65 & 0.92 & 0.34 & 0.64 & 0.93 \\
    Mahalanobis \cite{lee-2018} & 0.13 & 0.56 & 0.96 & 0.46 & 0.65 & 0.94 & 0.36 & 0.66 & 0.92 \\
    k-NN \cite{sun_out--distribution_2022} & 0.11 & 0.50 & 1.00 & 0.45 & 0.50 & 1.00 & 0.37 & 0.50 & 1.00 \\
    \hline
    PaRCE (Ours) & \textbf{0.47} & \textbf{0.80} & 0.78 & \textbf{0.73} & \textbf{0.83} & \textbf{0.74} & \textbf{0.47} & \textbf{0.73} & \textbf{0.82} \\
    \hline
    \end{tabular}
    }
    \caption{Comparison of measures of overall model competency for the lunar dataset with modified image properties.}
    \label{tab:overall-lunar-2}
\end{table*}

\begin{table*}[h!]
    \centering
    \renewcommand{\arraystretch}{1.2} 
    \resizebox{0.8\textwidth}{!}{%
    \begin{tabular}{|c|rrr|rrr|rrr|}
    \hline
    \multirow{2}{*}{Method}  & \multicolumn{3}{c|}{High vs. Medium} & \multicolumn{3}{c|}{High vs. Low} & \multicolumn{3}{c|}{Medium vs. Low} \\
    \cline{2-4} \cline{5-7} \cline{8-10}
     & Dist. $\uparrow$ & AUROC $\uparrow$ & FPR $\downarrow$ & Dist. $\uparrow$ & AUROC $\uparrow$ & FPR $\downarrow$ & Dist. $\uparrow$ & AUROC $\uparrow$ & FPR $\downarrow$ \\
    \hline
    \renewcommand{\arraystretch}{1} 
    Softmax & 0.34 & 0.71 & 0.91 & 0.78 & 0.80 & 0.84 & 0.49 & 0.76 & 0.83 \\
    Temperature \cite{guo} & 0.34 & 0.71 & 0.91 & 0.78 & 0.80 & 0.84 & 0.49 & 0.76 & 0.83 \\
    MC Dropout \cite{dropout} & 0.34 & 0.70 & 0.90 & 0.72 & 0.77 & 0.83 & 0.38 & 0.70 & 0.82 \\
    Ensemble \cite{lakshminarayanan_simple_2017} & 0.34 & 0.70 & 0.83 & 0.54 & 0.73 & 0.74 & 0.20 & 0.63 & 0.77 \\
    Energy \cite{liu_energy-based_2020} & 0.34 & 0.71 & 0.91 & 0.78 & 0.80 & 0.84 & 0.49 & 0.76 & 0.83 \\
    ODIN \cite{liang_enhancing_2020} & 0.02 & 0.49 & 1.00 & 0.27 & 0.55 & 1.00 & 0.28 & 0.58 & 1.00 \\
    OpenMax \cite{openmax} & 0.29 & 0.66 & 0.89 & 0.74 & 0.75 & 0.85 & 0.46 & 0.72 & 0.85 \\
    DICE \cite{dice} & 0.32 & 0.70 & 0.91 & 0.79 & 0.80 & 0.84 & \textbf{0.54} & \textbf{0.77} & 0.83 \\
    KL-Matching \cite{kl_matching} & 0.34 & 0.71 & 0.88 & 0.80 & 0.81 & 0.81 & 0.53 & \textbf{0.77} & 0.82 \\
    Mahalanobis \cite{lee-2018} & 0.00 & 0.41 & 0.98 & 0.00 & 0.31 & 0.99 & 0.00 & 0.30 & 0.99 \\
    k-NN \cite{sun_out--distribution_2022} & 0.00 & 0.34 & 0.99 & 0.00 & 0.23 & 1.00 & 0.00 & 0.24 & 0.99 \\
    \hline
    PaRCE (Ours) & \textbf{0.49} & \textbf{0.81} & \textbf{0.79} & \textbf{0.83} & \textbf{0.87} & \textbf{0.68} & 0.36 & \textbf{0.77} & \textbf{0.72} \\
    \hline
    \end{tabular}
    }
    \caption{Comparison of measures of overall model competency for the speed dataset with modified image properties.}
    \label{tab:overall-speed-2}
\end{table*}

\begin{table*}[h!]
    \centering
    \renewcommand{\arraystretch}{1.2} 
    \resizebox{0.8\textwidth}{!}{%
    \begin{tabular}{|c|rrr|rrr|rrr|}
    \hline
    \multirow{2}{*}{Method}  & \multicolumn{3}{c|}{High vs. Medium} & \multicolumn{3}{c|}{High vs. Low} & \multicolumn{3}{c|}{Medium vs. Low} \\
    \cline{2-4} \cline{5-7} \cline{8-10}
     & Dist. $\uparrow$ & AUROC $\uparrow$ & FPR $\downarrow$ & Dist. $\uparrow$ & AUROC $\uparrow$ & FPR $\downarrow$ & Dist. $\uparrow$ & AUROC $\uparrow$ & FPR $\downarrow$ \\
    \hline
    \renewcommand{\arraystretch}{1} 
    Softmax & 0.11 & 0.56 & 1.00 & 0.18 & 0.57 & 1.00 & 0.12 & 0.52 & 1.00 \\
    Temperature \cite{guo} & 0.38 & 0.52 & 1.00 & 0.55 & 0.54 & 1.00 & 0.25 & 0.53 & 0.94 \\
    MC Dropout \cite{dropout} & 0.14 & 0.58 & 0.95 & 0.44 & 0.61 & 0.94 & 0.31 & 0.55 & 0.94 \\
    Ensemble \cite{lakshminarayanan_simple_2017} & 0.30 & 0.66 & 1.00 & 0.76 & 0.70 & 1.00 & 0.50 & 0.58 & 1.00 \\
    Energy \cite{liu_energy-based_2020} & 0.11 & 0.56 & 1.00 & 0.18 & 0.57 & 1.00 & 0.12 & 0.51 & 1.00 \\
    ODIN \cite{liang_enhancing_2020} & 0.01 & 0.50 & 1.00 & 0.12 & 0.51 & 1.00 & 0.12 & 0.51 & 1.00 \\
    OpenMax \cite{openmax} & 0.11 & 0.56 & 0.96 & 0.13 & 0.55 & 0.96 & 0.03 & 0.49 & 0.98 \\
    DICE \cite{dice} & 0.05 & 0.51 & 0.96 & 0.14 & 0.48 & 1.00 & 0.12 & 0.45 & 1.00 \\
    KL-Matching \cite{kl_matching} & 0.25 & 0.58 & 1.00 & 0.68 & 0.62 & 1.00 & 0.64 & \textbf{0.60} & 0.94 \\
    Mahalanobis \cite{lee-2018} & 0.09 & 0.52 & 0.98 & 0.70 & 0.56 & 0.99 & 0.65 & 0.57 & 0.99 \\
    k-NN \cite{sun_out--distribution_2022} & 0.08 & 0.50 & 1.00 & 0.71 & 0.49 & 1.00 & \textbf{0.67} & 0.48 & 0.99 \\
    \hline
    PaRCE (Ours) & \textbf{0.59} & \textbf{0.85} & \textbf{0.81} & \textbf{0.88} & \textbf{0.87} & \textbf{0.77} & 0.37 & 0.57 & \textbf{0.93} \\
    \hline
    \end{tabular}
    }
    \caption{Comparison of measures of overall model competency for the park dataset with modified image properties.}
    \label{tab:overall-park-2}
\end{table*}

\textbf{Results:} The results are summarized in \cref{tab:overall-lunar-2,tab:overall-speed-2,tab:overall-park-2}. We also display the distribution of competency scores for high, medium, and low accuracy modifications in \cref{fig:box-eval2-lunar,fig:box-eval2-speed,fig:box-eval2-park} in Appendix \ref{sec:app-analysis}. 

The MSP and Temperature Scaling methods perform similarly to each other across all datasets. These methods are not insightful for the lunar or park datasets, assigning a near-maximum score to the majority of samples, regardless of model accuracy. For the speed dataset, these methods more clearly assign different distributions of scores to high, medium, and low accuracy image modifications, and their performance is more similar to other existing methods.

MC Dropout and Ensembling also perform similarly to one another. Their average scores tend to be similar across accuracy levels, but there is generally more variation for medium and low accuracy samples. However, the performance of these methods varies significantly across datasets.

For the classification-based OOD detection methods, there tends to be little variation in score distributions across high, medium, and low accuracy image modifications. They all generate more reasonable score distributions for the speed dataset, but they generally do not outperform previously mentioned methods, even for this dataset.

Looking at distance-based OOD detection methods, KL-matching generally assigns lower scores to low accuracy modifications than to high and medium accuracy ones, indicating some utility of this method. k-NN assigns lower scores on average for the low accuracy images in the lunar and park datasets but higher scores for the speed dataset. The Mahalanobis method assigns similar scores for high, medium, and low accuracy images across all datasets.

Our method performs the best across nearly all metrics for the modified datasets--best distinguishing between high, medium, and low accuracy image modifications. Again, other than the MSP, our method is the only probabilistic one, making it the most intuitive. It tends to assign scores close to one to images with high accuracy modifications, scores close to zero for low accuracy modifications, and scores in between zero and one for medium accuracy.

\subsection{Regional Competency Images}
\label{subsec:results-regional}

We compare our regional competency images to various existing methods for anomaly detection and localization. In particular, we compare our approach to GANomaly \cite{ganomaly}, DRAEM \cite{zavrtanik_draem_2021}, FastFlow \cite{fastflow}, PaDiM \cite{del_bimbo_padim_2021}, PatchCore \cite{patchcore}, Reverse Distillation \cite{reverse}, and Student-Teacher Feature Pyramid Matching \cite{stfpm}. These are several of the state-of-the-art anomaly detection algorithms that generate anomaly maps similar to our regional competency maps. All methods were implemented with the help of the Anomalib library \cite{akcay2022anomalib}.

\subsubsection{Evaluation: Anomalous Regions of Images}

\begin{table*}[h!]
    \centering
    \renewcommand{\arraystretch}{1.2} 
    \resizebox{0.75\textwidth}{!}{%
    \begin{tabular}{|c|r|rrr|rrr|rrr|}
    \hline
    \multirow{2}{*}{Method} & \multirow{2}{*}{\shortstack{Computation \\ Time (sec) $\downarrow$}} & \multicolumn{3}{c|}{ID All vs. OOD Unfamiliar} & \multicolumn{3}{c|}{OOD Familiar vs. OOD Unfamiliar} \\
    \cline{3-5} \cline{6-8}
     & & Dist. $\uparrow$ & AUROC $\uparrow$ & FPR $\downarrow$ & Dist. $\uparrow$ & AUROC $\uparrow$ & FPR $\downarrow$ \\
    \hline
    \renewcommand{\arraystretch}{1} 
    GANomaly \cite{ganomaly} & 0.074 & 0.11 & 0.58 & 0.91 & 0.11 & 0.56 & 0.92 \\
    DRAEM \cite{zavrtanik_draem_2021} & 0.197 & 0.36 & 0.68 & 0.97 & 0.46 & 0.69 & 0.97 \\
    FastFlow \cite{fastflow} & 0.036 & 0.84 & 0.97 & 0.15 & 0.74 & 0.96 & 0.17 \\
    PaDiM \cite{del_bimbo_padim_2021} & 0.026 & 0.89 & \textbf{0.98} & 0.07 & 0.80 & \textbf{0.98} & 0.09 \\
    PatchCore \cite{patchcore} & 0.640 & \textbf{0.94} & 0.59 & 0.83 & 0.78 & 0.57 & 0.86 \\
    Reverse Distillation \cite{reverse} & 0.083 & 0.85 & 0.95 & 0.10 & 0.80 & 0.95 & 0.11 \\
    Student-Teacher \cite{stfpm} & \textbf{0.014} & 0.79 & 0.95 & 0.33 & 0.75 & 0.95 & 0.32 \\
    \hline
    PaRCE (Ours) & 0.122 & 0.89 & \textbf{0.98} & \textbf{0.06} & \textbf{0.87} & \textbf{0.98} & \textbf{0.07} \\
    \hline
    \end{tabular}
    }
    \caption{Comparison of measures of regional model competency for the lunar dataset with anomalous regions.}
    \label{tab:regional-lunar}
\end{table*}

\begin{table*}[h!]
    \centering
    \renewcommand{\arraystretch}{1.2} 
    \resizebox{0.75\textwidth}{!}{%
    \begin{tabular}{|c|r|rrr|rrr|rrr|}
    \hline
    \multirow{2}{*}{Method} & \multirow{2}{*}{\shortstack{Computation \\ Time (sec) $\downarrow$}} & \multicolumn{3}{c|}{ID All vs. OOD Unfamiliar} & \multicolumn{3}{c|}{OOD Familiar vs. OOD Unfamiliar} \\
    \cline{3-5} \cline{6-8}
     & & Dist. $\uparrow$ & AUROC $\uparrow$ & FPR $\downarrow$ & Dist. $\uparrow$ & AUROC $\uparrow$ & FPR $\downarrow$ \\
    \hline
    \renewcommand{\arraystretch}{1} 
    GANomaly \cite{ganomaly} & 0.075 & 0.15 & 0.55 & 0.80 & 0.08 & 0.55 & 0.80 \\
    DRAEM \cite{zavrtanik_draem_2021} & 0.118 & 0.22 & 0.62 & 0.85 & 0.19 & 0.62 & 0.85 \\
    FastFlow \cite{fastflow} & 0.035 & 0.43 & 0.77 & 0.78 & 0.34 & 0.77 & 0.78 \\
    PaDiM \cite{del_bimbo_padim_2021} & 0.025 & 0.58 & 0.86 & \textbf{0.42} & 0.43 & 0.86 & \textbf{0.44} \\
    PatchCore \cite{patchcore} & 0.870 & 0.45 & 0.50 & 1.00 & 0.33 & 0.50 & 1.00 \\
    Reverse Distillation \cite{reverse} & 0.068 & 0.22 & 0.62 & 0.79 & 0.16 & 0.62 & 0.79 \\
    Student-Teacher \cite{stfpm} & \textbf{0.014} & 0.32 & 0.69 & 0.72 & 0.18 & 0.69 & 0.72 \\
    \hline
    PaRCE (Ours) & 0.145 & \textbf{0.75} & \textbf{0.89} & 0.54 & \textbf{0.52} & \textbf{0.89} & 0.54 \\
    \hline
    \end{tabular}
    }
    \caption{Comparison of measures of regional model competency for the speed dataset with anomalous regions.}
    \label{tab:regional-speed}
\end{table*}

\begin{table*}[h!]
    \centering
    \renewcommand{\arraystretch}{1.2} 
    \resizebox{0.75\textwidth}{!}{%
    \begin{tabular}{|c|r|rrr|rrr|rrr|}
    \hline
    \multirow{2}{*}{Method} & \multirow{2}{*}{\shortstack{Computation \\ Time (sec) $\downarrow$}} & \multicolumn{3}{c|}{ID All vs. OOD Unfamiliar} & \multicolumn{3}{c|}{OOD Familiar vs. OOD Unfamiliar} \\
    \cline{3-5} \cline{6-8}
     & & Dist. $\uparrow$ & AUROC $\uparrow$ & FPR $\downarrow$ & Dist. $\uparrow$ & AUROC $\uparrow$ & FPR $\downarrow$ \\
    \hline
    \renewcommand{\arraystretch}{1} 
    GANomaly \cite{ganomaly} & 0.036 & 0.06 & 0.53 & 0.93 & 0.07 & 0.53 & 0.93 \\
    DRAEM \cite{zavrtanik_draem_2021} & 0.120 & 0.09 & 0.51 & 0.95 & 0.17 & 0.51 & 0.95 \\
    FastFlow \cite{fastflow} & 0.034 & 0.52 & 0.83 & 0.72 & 0.28 & 0.83 & 0.72 \\
    PaDiM \cite{del_bimbo_padim_2021} & 0.025 & 0.54 & 0.84 & 0.73 & 0.28 & 0.84 & 0.73 \\
    PatchCore \cite{patchcore} & 0.937 & \textbf{0.90} & 0.51 & 0.98 & \textbf{0.36} & 0.51 & 0.98 \\
    Reverse Distillation \cite{reverse} & 0.070 & 0.45 & 0.77 & 0.72 & 0.17 & 0.76 & 0.72 \\
    Student-Teacher \cite{stfpm} & \textbf{0.014} & 0.67 & \textbf{0.92} & \textbf{0.39} & 0.34 & \textbf{0.92} & \textbf{0.39} \\
    \hline
    PaRCE (Ours) & 0.296 & 0.45 & 0.74 & 0.94 & 0.16 & 0.74 & 0.94 \\
    \hline
    \end{tabular}
    }
    \caption{Comparison of measures of regional model competency for the park dataset with anomalous regions.}
    \label{tab:regional-park}
\end{table*}

\textbf{Datasets:} We conduct analysis using the three datasets discussed in \cref{subsubsec:results-overall-1}. These datasets allow us to assess the ability of methods to identify unfamiliar objects, detect regions associated with unseen classes, and recognize unexplored areas in an environment. See \cref{sec:app-datasets} for details.


\textbf{Metrics:} We evaluate methods based on their computation time and their ability to distinguish between familiar regions (in both ID and OOD images) and unfamiliar regions (in OOD images). Familiar regions are all of the pixels that occupy image structures that exist in the training set, and unfamiliar pixels are those that occupy structures that were not present during training. To evaluate the ability to distinguish between sets of samples using each mapping method, we consider the same metrics discussed in \cref{subsubsec:results-overall-1}.

\textbf{Results:} The results are summarized in \cref{tab:regional-lunar,tab:regional-speed,tab:regional-park}. We display the distribution of competency scores for pixels in ID images, familiar pixels in OOD images, and unfamiliar pixels in OOD images in \cref{fig:box-eval3-lunar,fig:box-eval3-speed,fig:box-eval3-park} in Appendix \ref{sec:app-analysis}. Examples of the regional competency images are shown in \cref{fig:regional-lunar,fig:regional-speed,fig:regional-park} in Appendix \ref{sec:app-analysis}.

GANomaly, DRAEM, and Reverse Distillation generally perform worse than competing methods and never outperform other methods because they tend to assign similar scores to both familiar and unfamiliar pixels. 

FastFlow and PaDiM tend to perform comparably to one another, generating similar score distributions and competency images. They generally perform better than the first three mentioned methods. Quantitatively, PaDiM tends to perform a bit better that FastFlow. It is one of the methods with the highest AUROC scores for the lunar dataset and achieves the lowest FPRs for the speed limit signs dataset.

PatchCore is by far the slowest method, requiring around one second to obtain an output for a single image. It tends to obtain high KS distances, achieving the highest distances for the park dataset and the highest distance between ID and unfamiliar OOD distributions for the lunar dataset. However, it also tends to obtain very high (and often the highest) FPRs due to the large range of scores for familiar pixels.

Student-Teacher Feature Pyramid Matching is consistently the fastest method, maintaining the same low average computation time across all datasets. It does not perform particularly well for the lunar or speed datasets but performs better across a number of metrics for the park dataset, achieving the highest AUROC scores and lowest FPRs. 

Our method, PaRCE, scores best across the majority of metrics for the lunar and speed datasets. For the lunar dataset, it achieves the highest or second highest KS distances, highest AUROC scores, and lowest FPRs. For the speed dataset, it achieves the highest KS distances, highest AUROCs, and second lowest FPRs. Along with DRAEM, it is one of only two methods with a consistent score range. Unlike other methods, it outputs intuitive probability scores that tend to be around one for familiar pixels and below one for unfamiliar pixels. Despite its success on the lunar and speed datasets, like most other methods, it performs quite poorly for the park dataset. This reduction in performance is likely due to limitations of the simple Felzenszwalb algorithm \cite{felzenszwalb_efficient_2004} in segmenting these more complex images.

\section{Conclusions}
\label{sec:conclusion}

In this work, we present a probabilistic and reconstruction-based competency estimation (PaRCE) method that captures predictive uncertainty in CNN-based perception models by estimating prediction accuracy. Through comparisons with various UQ and OOD detection techniques, we demonstrate that PaRCE effectively distinguishes between correctly classified, misclassified, and OOD samples. It also shows robust performance in identifying visual image modifications that correspond to high, medium, and low prediction accuracy. Additionally, we extend PaRCE to generate regional competency maps, which reflect model confidence for specific regions within an image. Our comparisons with existing anomaly localization techniques indicate that PaRCE often outperforms other methods in distinguishing familiar from unfamiliar pixels within an image.

Despite these strengths, we identified a dataset for which PaRCE, alongside other anomaly localization methods, performs suboptimally. For PaRCE in particular, this decrease in performance could stem from challenges in segmenting images from complex, heterogeneous environments. In our work, images are segmented using the graph-based segmentation algorithm developed by Felzenszwalb \cite{felzenszwalb_efficient_2004}. This algorithm is not trainable and has not been tailored towards the specific datasets or tasks we consider. It is useful because it is fast and can theoretically work with any dataset, but it often selects non-intuitive segmentations, particularly for the park dataset.  Future work could explore domain-specific segmentation methods to enhance performance. 

To further improve performance, one could consider the use of different reconstruction loss functions or ensembles of autoencoders. Another promising direction would involve integrating competency scores into decision-making frameworks that benefit from risk assessment, particularly for autonomous systems or those that rely on human experts sparingly. Finally, while we evaluated competency estimation methods using autonomous navigation datasets, future work should evaluate methods in other domains, such as medical imaging and industrial inspection, where the need for robust and trustworthy models is equally critical.

{
    \small
    \bibliographystyle{ieeenat_fullname}
    \bibliography{main}
}


\onecolumn
\appendix

\section{Data, Model, \& Code Sources}
\label{sec:code}

The \href{https://huggingface.co/datasets/sarapohland/lunar-navigation}{lunar}, \href{https://huggingface.co/datasets/sarapohland/speed-limit-signs}{speed}, and \href{https://huggingface.co/datasets/sarapohland/park-navigation}{park} datasets we use for evaluation, as well as the trained \href{https://huggingface.co/sarapohland/lunar-navigation}{lunar}, \href{https://huggingface.co/sarapohland/speed-limit-signs}{speed}, and \href{https://huggingface.co/sarapohland/park-navigation}{park} models, are provided through \href{https://huggingface.co/collections/sarapohland/parce-67a1479f5efa4fab15372a64}{HuggingFace}. For each of the models, the model architecture is described in a JSON file called layers.json and training parameters are provided in a configuration file called train.config. All of our code is available in our GitHub \href{https://github.com/sarapohland/parce.git}{repository}, and instructions for reproducing the results in this paper are provided in the \href{https://github.com/sarapohland/parce/blob/main/README.md}{README}.


\section{Datasets for Evaluation}
\label{sec:app-datasets}


Our work is largely motivated by our experiences with CNN-based perception systems for the autonomous navigation of robots in outdoor environments. We noticed that small changes in the vehicle's environment could lead to very unreasonable predictions and sought to better quantify perception model confidence under such changes. 

While there are a large number of existing datasets focused on domain shift and generalization (e.g., ImageNet variants, PACS, Office-31/Office-Home, DomainNet, VisDA, VLCS, WILDS, Terra Incognita, and Digits-DG), these datasets are generally focused on recognizing animals, objects, or digits and are not particularly relevant to robot navigation. Within the area of anomaly detection, there are a number of datasets focused on industrial inspection (e.g., MVTec) or medical anomalies (e.g., BRATS and CheXpert), but, again, these datasets are irrelevant for robot navigation. While there are also a number of existing datasets for indoor navigation and autonomous driving, there are few focused on outdoor robot navigation. These datasets (NCLT, Oxford Offroad Radar, Symphony Seasons SFU Mountain and Montmorency, UTIAS, and BotanicGarden) vary in popularity and would need to be adapted to be useful in evaluating competency estimation of image classifiers.

Focusing on perception models for navigation, we observe two main causes that tend to lead to a reduction in prediction accuracy: (1) particular parts of the image/regions in the environment are unfamiliar to the perception model and (2) holistic image properties are causing a change in the expected model prediction. To address the first cause, we consider three datasets whose OOD set contains either (a) unfamiliar objects, (b) a class that is not part of the training set, or (c) regions outside of the environment in which the model was trained. We provide additional details on these datasets in Section \ref{subsec:app-spatial}. To address the second cause, we considered the impact of seven image properties (saturation, contrast, brightness, sharpness, blurriness, noise, and pixelation) on the prediction accuracy of various perception navigation models and developed datasets by varying the properties most salient to accuracy. We discuss these datasets in more detail in Section \ref{subsec:app-modified}.

\subsection{Generation of Datasets with Spatial Anomalies} \label{subsec:app-spatial}

We conduct analysis across three unique datasets that contain manually identified OOD samples. The first dataset was obtained from a lunar environment, in which the training data contains images from an uninhabited moon and the test data contains images from a habited moon. While the training images only contain the lunar surface, the sky, and shadows, the test images additionally contain astronauts and human-made structures. This dataset enables us to assess the ability of overall competency approaches to detect images with unfamiliar objects and regional competency methods to identify these objects. Column 1 of \cref{fig:lunar} displays example in-distribution images from this dataset, and column 2 shows example OOD images.

The second dataset contains speed limit signs in Germany \cite{gtsdb}. While the training dataset is composed of common speed limit signs (30 km/hr and higher), the test dataset set also contains signs with an uncommon speed limit (20 km/hr). This dataset assesses the ability of overall competency methods to identify images associated with unfamiliar class labels and the ability of regional competency estimators to detect regions associated with previously unseen classes. Column 1 of \cref{fig:speed} displays example in-distribution images from this dataset, and column 2 shows example OOD images.

The third dataset contains images from regions in a park. While the training dataset only contains images from forested and grassy regions of the park, the test dataset additionally includes images from around the park's pavilion. This dataset allows for the assessment of the ability of overall competency methods to identify images from unexplored regions in an environment and for regional approaches to further identify these regions within a given image. Column 1 of \cref{fig:park} displays example in-distribution images from this dataset, and column 2 shows example OOD images.

\subsection{Generation of Datasets with Non-Spatial Anomalies} \label{subsec:app-modified}

\begin{figure*}[h!]
  \centering
  \includegraphics[width=0.7\linewidth]{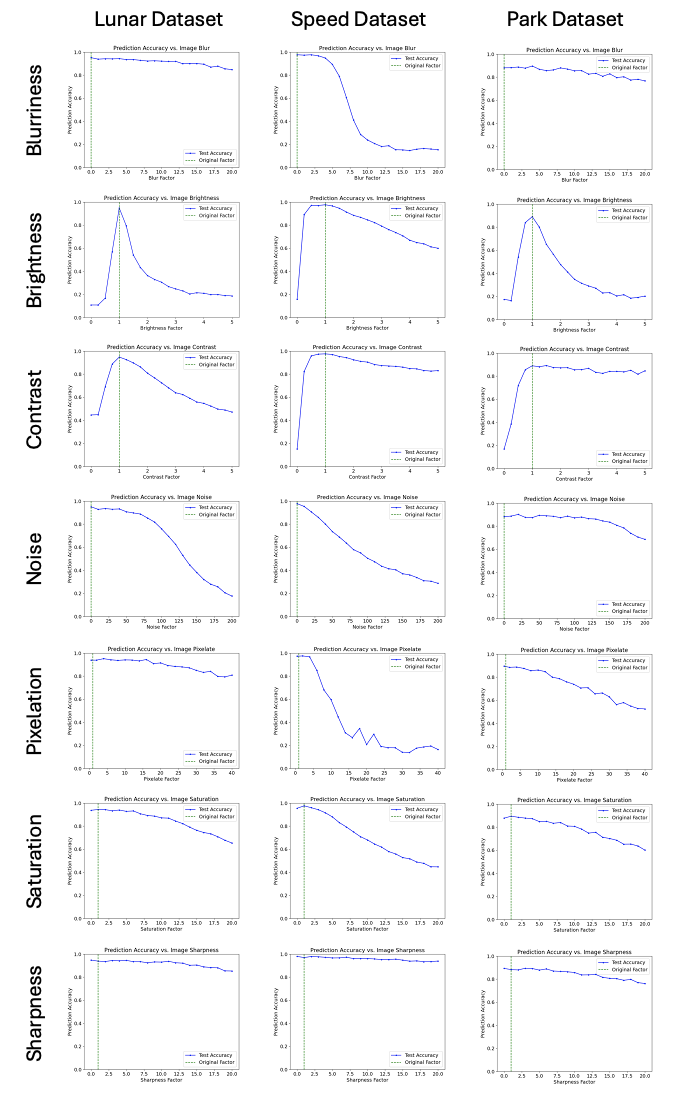}
  \caption{Plots of in-distribution test accuracy across 21 factors of blurriness, brightness, contrast, noise, pixelation, saturation, and sharpness for classification models trained on the lunar, speed, and park datasets. The blue line indicates test accuracy, and the dotted green line shows the original/default factor, which is either zero or one for all image properties.}
  \label{fig:image-properties}
\end{figure*}

While many existing works in OOD detection and anomaly localization focus on spatial anomalies--where specific regions in the image are anomalous--we are also interested in the ability of existing methods to identify non-regional anomalies that reduce model confidence. We consider how prediction accuracy is impacted by seven image properties: blurriness, brightness, contrast, noise, pixelation, saturation, and sharpness. 
Figure \ref{fig:image-properties} displays in-distribution test accuracy versus 21 values of each image property for the three datasets described in Section \ref{subsec:app-spatial}. From these results, we noticed that blurriness and sharpness often do not result in a significant reduction in prediction accuracy, so we chose to exclude these two properties. These result also help us select and interesting range of property values for each dataset, which are summarized in Table \ref{tab:property values}.

\begin{figure*}[h!]
  \centering
  \includegraphics[width=0.83\linewidth]{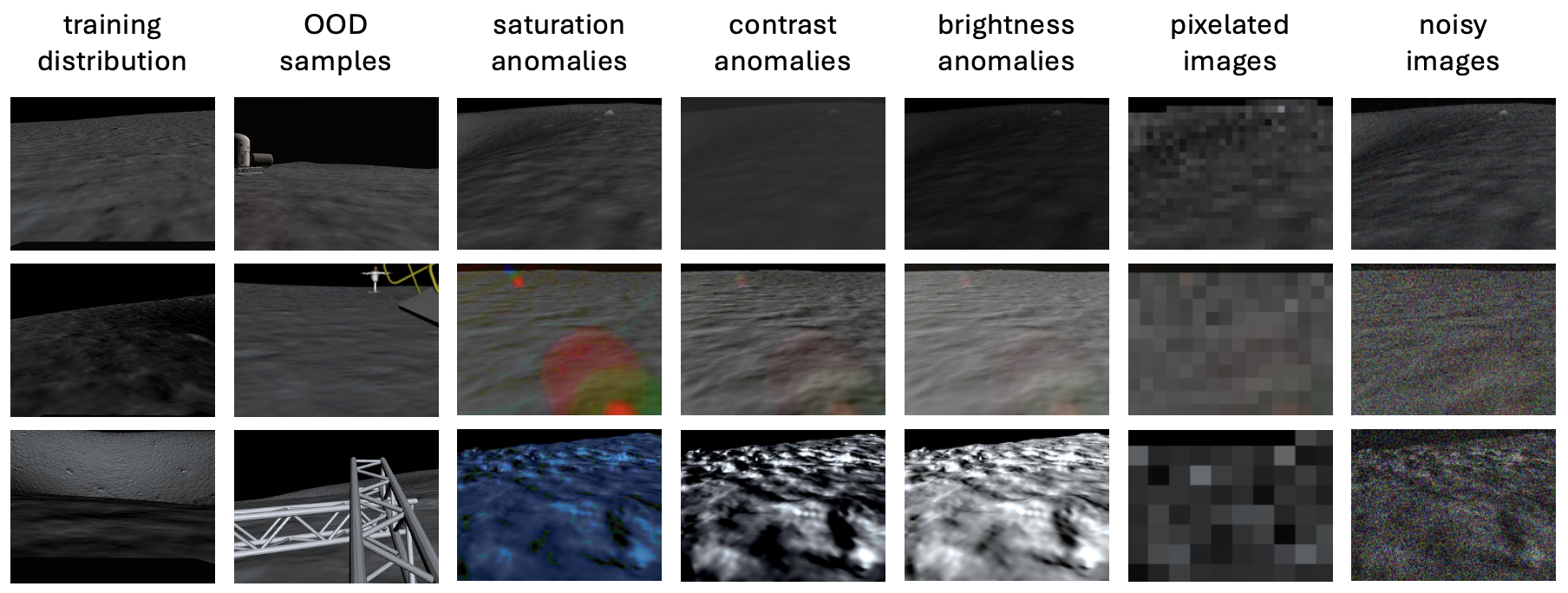}
  \caption{Example images from the \textit{lunar} dataset. Column 1: test images drawn from the same distribution as the training images (uninhabited moon). Column 2: test images with unfamiliar obstacles drawn from a new distribution (habited moon). Column 3: test images with varied saturation levels generated from in-distribution images. Column 4: test images with varied contrast levels generated from in-distribution images. Column 5: test images with varied brightness levels generated from in-distribution images. Column 6: in-distribution images that have been pixelated. Column 7: in-distribution images with additive noise.}
  \label{fig:lunar}
\end{figure*}

\begin{figure*}[h!]
  \centering
  \includegraphics[width=0.83\linewidth]{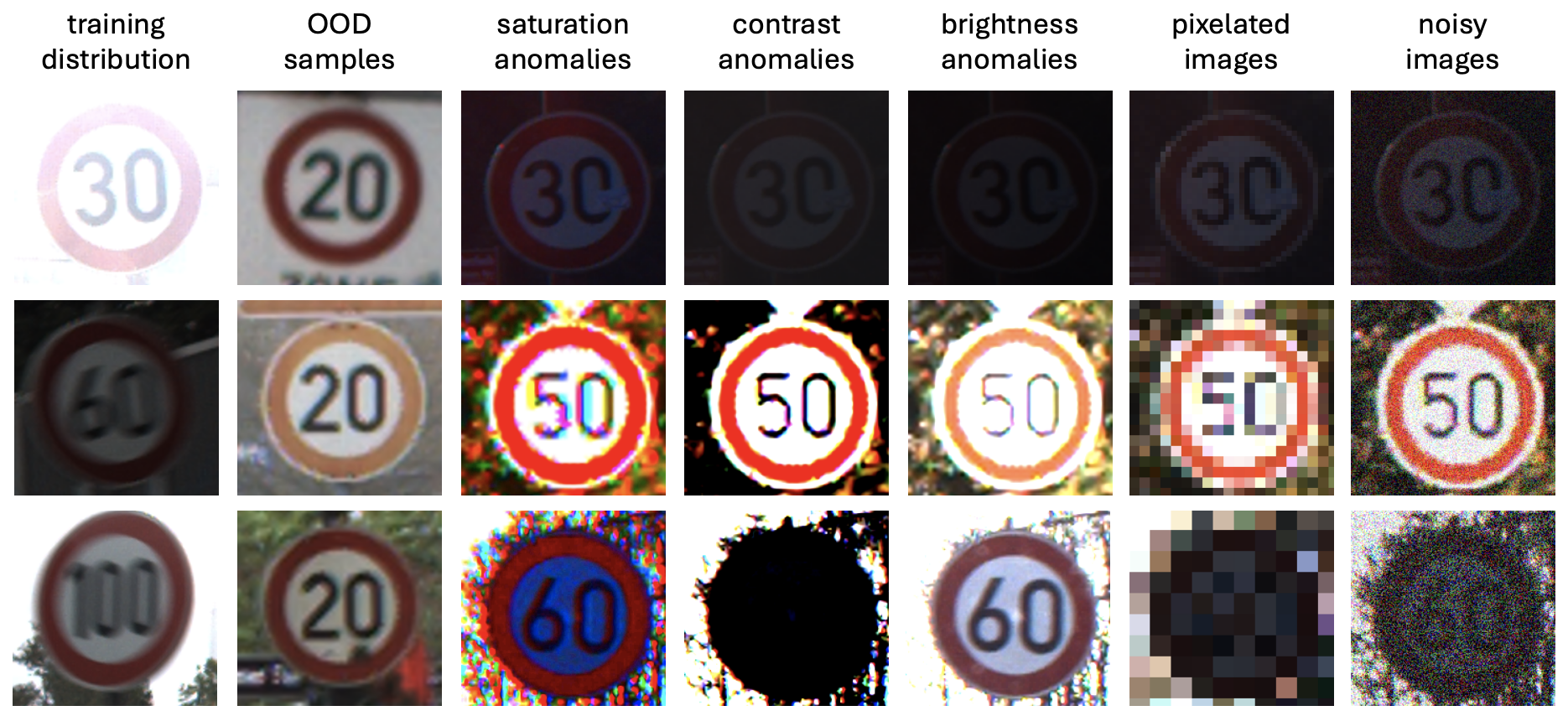}
  \caption{Example images from the \textit{speed limit signs} dataset. Column 1: test images drawn from the same distribution as the training images (common speed limit signs). Column 2: test images with unfamiliar class labels drawn from a new distribution (uncommon speed limit signs). Column 3: test images with varied saturation levels generated from in-distribution images. Column 4: test images with varied contrast levels generated from in-distribution images. Column 5: test images with varied brightness levels generated from in-distribution images. Column 6: in-distribution images that have been pixelated. Column 7: in-distribution images with additive noise.}
  \label{fig:speed}
\end{figure*}

\begin{figure*}[h!]
  \centering
  \includegraphics[width=0.83\linewidth]{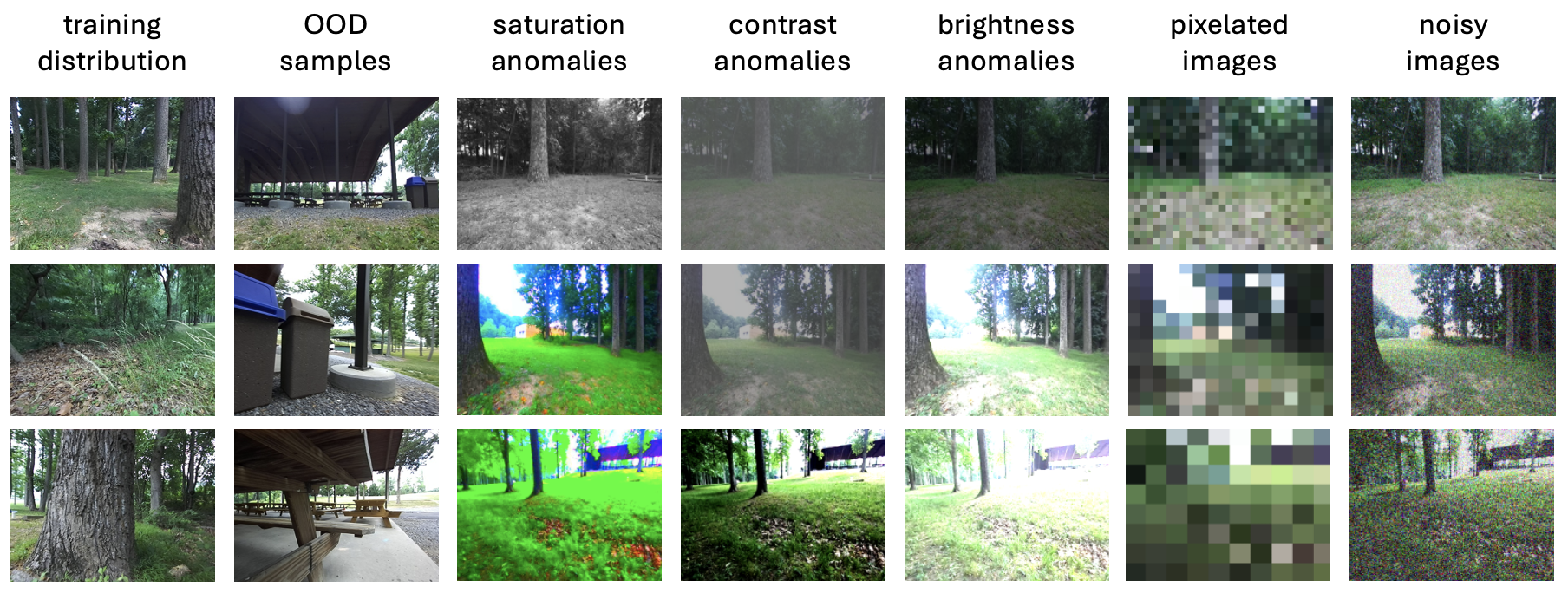}
  \caption{Example images from the \textit{outdoor park} dataset. Column 1: test images drawn from the same distribution as the training images (forested/grassy regions). Column 2: test images associated with unexplored regions drawn from a new distribution (regions around pavilion). Column 3: test images with varied saturation levels generated from in-distribution images. Column 4: test images with varied contrast levels generated from in-distribution images. Column 5: test images with varied brightness levels generated from in-distribution images. Column 6: in-distribution images that have been pixelated. Column 7: in-distribution images with additive noise.}
  \label{fig:park}
\end{figure*}

For each of the three mentioned datasets, we generate new data from the in-distribution samples using the factors in Table \ref{tab:property values}. Each image modification results in either high (0.9 to 1.0), medium (0.5 to 0.9), or low (0.0 to 0.5) prediction accuracy. We are interested in whether overall competency estimation approaches can distinguish between visual image modifications that result in varying levels of accuracy. Column 1 of \cref{fig:lunar,fig:speed,fig:park} displays examples of unmodified images from each of the original datasets, and columns 3-7 of these figures show examples of generated images.


\begin{table}[h!]
    \centering
    \caption{Image property factors used for evaluation of competency estimation with modified datasets.}
    \renewcommand{\arraystretch}{1.2}
    \begin{tabular}{|c|c|c|c|}
        \hline
        Image Property & Lunar Dataset & Speed Dataset & Park Dataset \\
        \hline
        Brightness & 0.5, 0.75, 1.25, 1.5, 2.0, 3.0 & 0.25, 0.5, 2.0, 3.0, 4.0, 5.0 & 0.25, 0.5, 1.5, 2.0, 2.5, 3.0 \\
        Contrast & 0.25, 0.5, 1.5, 2.0, 3.0, 4.0 & 0.25, 0.5, 2.0, 3.0, 4.0, 5.0 & 0.25, 0.5, 0.75, 1.5, 1.75, 2.0 \\
        Noise & 20, 40, 60, 80, 100, 120 & 20, 40, 60, 80, 100, 120 & 20, 40, 60, 80, 100, 120 \\
        Pixelation & 10, 20, 25, 30, 35, 40 & 4, 8, 12, 16, 20, 24 & 10, 20, 25, 30, 35, 40 \\
        Saturation & 0, 5, 10, 15, 20, 25 & 0, 2, 4, 8, 12, 16 & 0, 0.5, 2.5, 5, 7.5, 10 \\
        \hline
    \end{tabular}
    \label{tab:property values}
\end{table}

\section{Additional Details on PaRCE Score}
\label{sec:app-definition}

\begin{figure*}[h!]
  \centering
  \begin{subfigure}{0.33\linewidth}
    \includegraphics[width=0.9\linewidth]{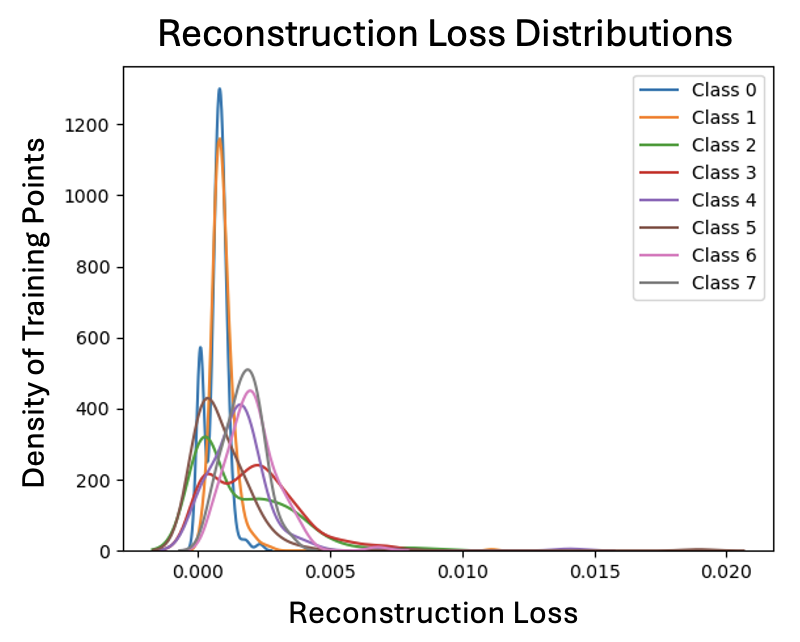}
    \caption{Lunar dataset}
    \label{fig:reco-lunar}
  \end{subfigure}
  \hfill
  \begin{subfigure}{0.33\linewidth}
    \includegraphics[width=0.9\linewidth]{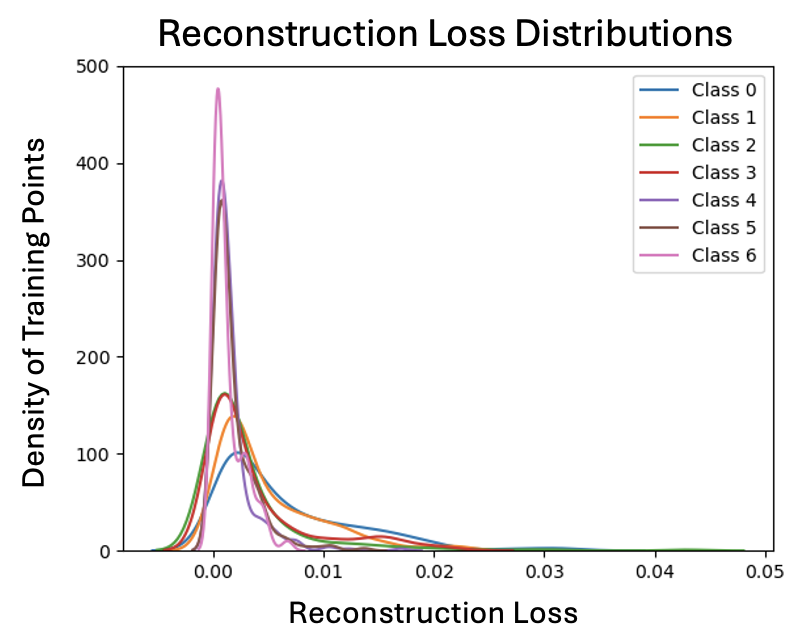}
    \caption{Speed dataset}
    \label{fig:reco-speed}
  \end{subfigure}
  \hfill
  \begin{subfigure}{0.33\linewidth}
    \includegraphics[width=0.9\linewidth]{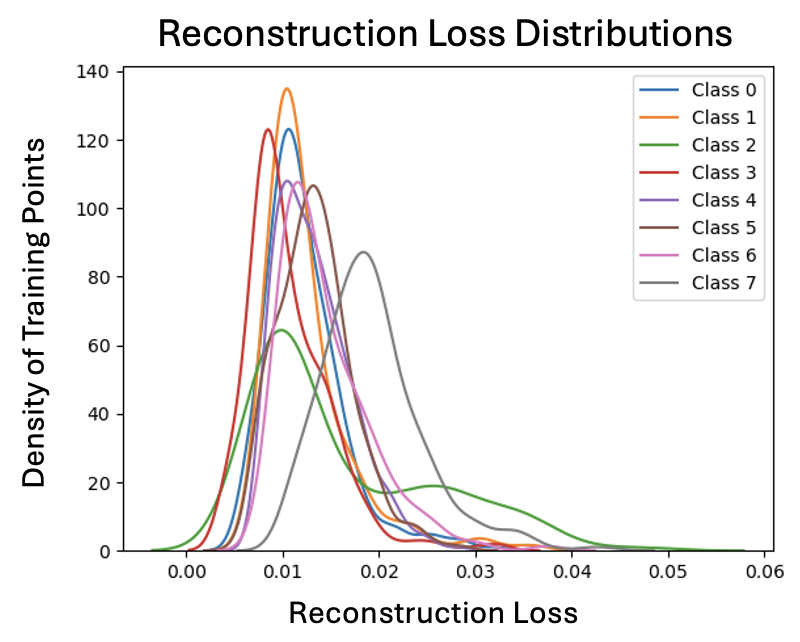}
    \caption{Park dataset}
    \label{fig:reco-park}
  \end{subfigure}
  \caption{True distribution of reconstruction losses obtained by the trained image reconstruction model for images in the in-distribution holdout set of the (a) lunar, (b) speed limit signs, and (c) outdoor park datasets across class labels.}
  \label{fig:reco}
\end{figure*}

\begin{figure*}[h!]
  \centering
  \begin{subfigure}{0.33\linewidth}
    \includegraphics[width=0.9\linewidth]{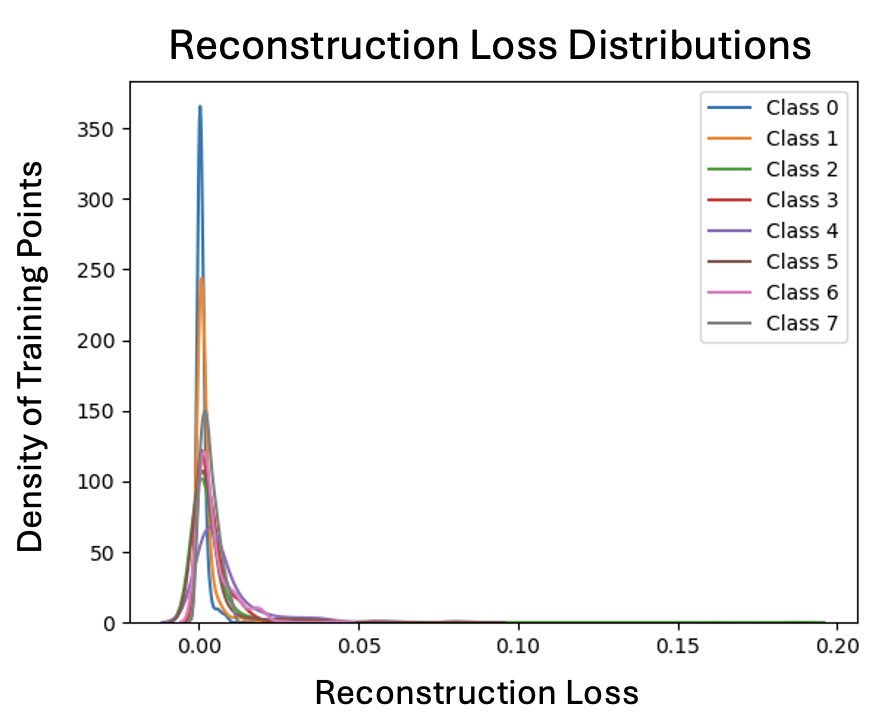}
    \caption{Lunar dataset}
    \label{fig:inpaint-lunar}
  \end{subfigure}
  \hfill
  \begin{subfigure}{0.33\linewidth}
    \includegraphics[width=0.9\linewidth]{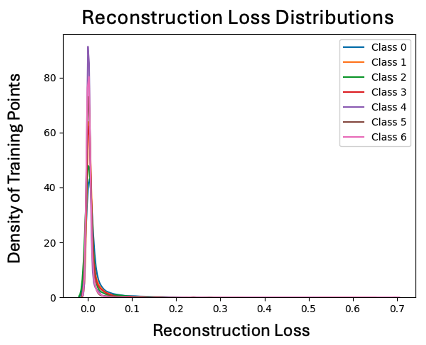}
    \caption{Speed dataset}
    \label{fig:inpaint-speed}
  \end{subfigure}
  \hfill
  \begin{subfigure}{0.33\linewidth}
    \includegraphics[width=0.9\linewidth]{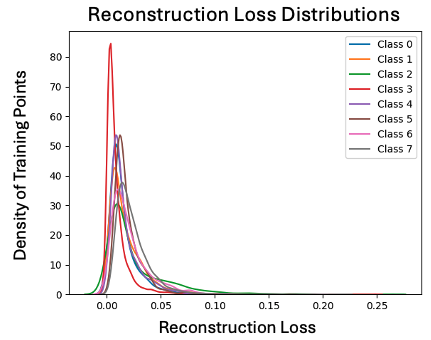}
    \caption{Park dataset}
    \label{fig:inpaint-park}
  \end{subfigure}
  \caption{True distribution of reconstruction losses obtained by the trained segment inpainting model for images in the in-distribution holdout set of the (a) lunar, (b) speed limit signs, and (c) outdoor park datasets across class labels.}
  \label{fig:inpaint}
\end{figure*}

To estimate the in-distribution probability of our overall PaRCE score, we design an autoencoder to reconstruct the input image. A holdout set of images, drawn from the same distribution as the training samples but not seen by the perception model, is used to estimate the distribution of the reconstruction loss for each class. The distributions for this set across each of the datasets described in \cref{sec:app-datasets} are displayed in \cref{fig:reco}. Similarly, to estimate the in-distribution probability of our regional PaRCE score, we design an autoencoder to inpaint missing segments of an image. A holdout set of segments is used to estimate the distributions of reconstruction loss, which are displayed in \cref{fig:inpaint} for each of the datasets.

The probability that an image, $\bm{X}$, is drawn from the same distribution as those in the training sample is given by \cref{eq:id-prob}. We estimate $P(D|\{f(\bm{X})=c\})$ as the probability that the reconstruction loss, $\ell(\bm{X})$, corresponding to image $\bm{X}$ aligns with the bottom $N\%$ of losses for the training distribution. For each class, we assume the reconstruction loss, $\mathcal{L}_c$, follows a Gaussian distribution with mean $\mu_c$ and standard deviation $\sigma_c$. Because $\mathcal{L}_c$ is a Gaussian random variable, $N$ corresponds to a z-score, $z$. We then estimate the class in-distribution probability, $P(D|\{f(\bm{X})=c\})$, in the following way:
\begin{align}
    \hat{P}(D|\{f(\bm{X})=c\})
    &= P\bigl(\{\mathcal{L}_c > \ell(\bm{X}) - (\mu_c + z\sigma_c)\}\bigr) \label{eq:deriv-1} \\
    &= 1 - P\bigl(\{\mathcal{L}_c \leq \ell(\bm{X}) - (\mu_c + z\sigma_c)\}\bigr) \label{eq:deriv-2} \\
    &= 1 - F_{\mathcal{L}_c\sim\mathcal{N}(\mu_c,\sigma_c)}\bigl(\ell(\bm{X}) - \mu_c - z\sigma_c\bigr) \label{eq:deriv-3} \\
     &= 1 - F_{Z\sim\mathcal{N}(0,1)}\left(\frac{\ell(\bm{X}) - 2\mu_c - z\sigma_c}{\sigma_c}\right) \label{eq:deriv-4} \\
     &=: 1 - \phi\left(\frac{\ell(\bm{X}) - 2\mu_c}{\sigma_c} - z\right). \label{eq:deriv-5}
\end{align}

\begin{figure*}[h!]
  \centering
  \begin{subfigure}{0.33\linewidth}
    \includegraphics[width=0.9\linewidth]{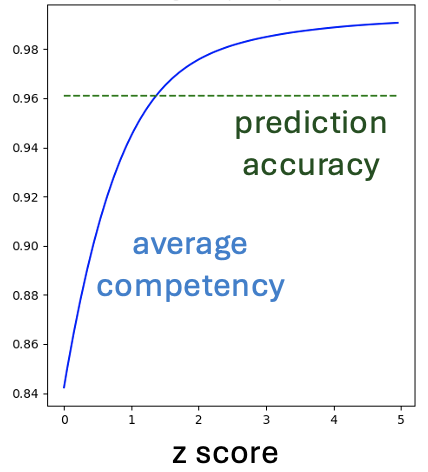}
    \caption{Lunar dataset}
    \label{fig:zscore-lunar}
  \end{subfigure}
  \hfill
  \begin{subfigure}{0.33\linewidth}
    \includegraphics[width=0.9\linewidth]{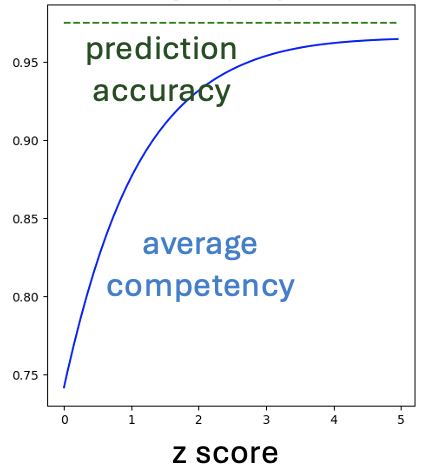}
    \caption{Speed dataset}
    \label{fig:zscore-speed}
  \end{subfigure}
  \hfill
  \begin{subfigure}{0.33\linewidth}
    \includegraphics[width=0.9\linewidth]{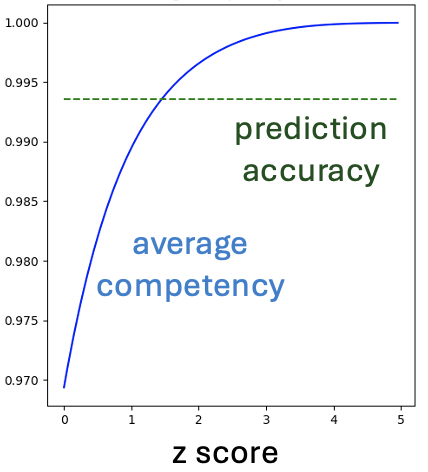}
    \caption{Park dataset}
    \label{fig:zscore-park}
  \end{subfigure}
  \caption{Plot of image prediction accuracy and average overall competency score for the in-distribution holdout set (defined in \cref{eq:calibration}) versus the z-score parameter (from \cref{eq:estimated-competency}) for the (a) lunar, (b) speed limit signs, and (c) outdoor park datasets.}
  \label{fig:zscore}
\end{figure*}

\begin{figure*}[h!]
  \centering
  \begin{subfigure}{0.33\linewidth}
    \includegraphics[width=0.9\linewidth]{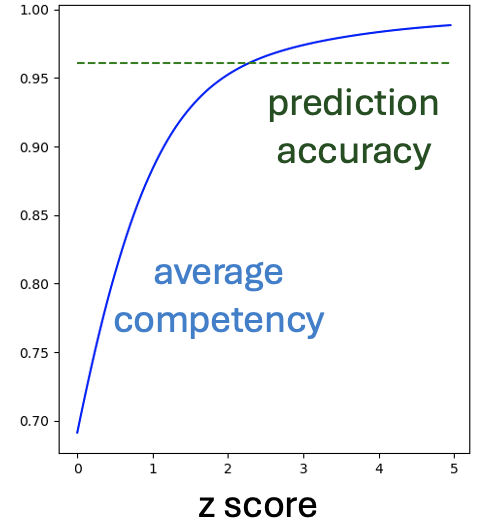}
    \caption{Lunar dataset}
    \label{fig:zscore-lunar-seg}
  \end{subfigure}
  \hfill
  \begin{subfigure}{0.33\linewidth}
    \includegraphics[width=0.9\linewidth]{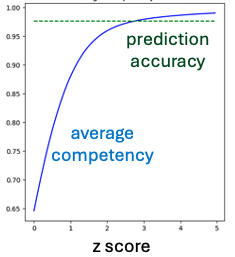}
    \caption{Speed dataset}
    \label{fig:zscore-speed-seg}
  \end{subfigure}
  \hfill
  \begin{subfigure}{0.33\linewidth}
    \includegraphics[width=0.9\linewidth]{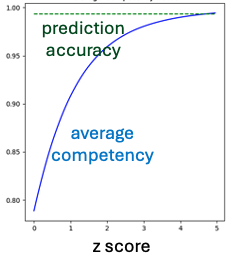}
    \caption{Park dataset}
    \label{fig:zscore-park-seg}
  \end{subfigure}
  \caption{Plot of segment prediction accuracy and average regional competency score for the in-distribution holdout set (defined in \cref{eq:calibration}) versus the z-score parameter (from \cref{eq:estimated-competency}) for the (a) lunar, (b) speed limit signs, and (c) outdoor park datasets.}
  \label{fig:zscore-seg}
\end{figure*}

\newpage
To calibrate our competency estimator, we select the z-score parameter such that the average competency score (left side of \cref{eq:calibration}) reflects the prediction accuracy (right side of \cref{eq:calibration}) for the in-distribution holdout set. \cref{fig:zscore} displays a plot of the prediction accuracy and average overall competency score for the holdout set across various z-score values for each of the three datasets. \cref{fig:zscore-seg} displays the prediction accuracy and average regional competency score. The z-score is chosen such that the absolute difference between the prediction accuracy and average competency score is minimized. 

\subsection{Overall Competency Score}
\label{subsec:app-definition-overall}

\newpage
\subsection{Regional Competency Image}
\label{subsec:app-definition-regional}

\section{Additional Analysis of Competency Scores}
\label{sec:app-analysis}


We evaluate both overall and regional competency estimation methods based on their computation time and their ability to distinguish between various types of data. For the overall competency estimation approaches, we assess their ability to distinguish between correctly classified, misclassified, and OOD samples. In addition, we evaluate their ability to distinguish between visual image modifications resulting in high, medium, and low prediction accuracy. (Refer to \cref{sec:app-datasets} for additional details on these data types.) Comparing regional competency estimation approaches, we consider the ability to distinguish between familiar pixels (from both in-distribution and OOD images) and unfamiliar pixels (from OOD samples). Familiar regions are all of the pixels that occupy image structures that exist in the training set, and unfamiliar pixels are those that occupy structures that were not present during training. To quantify the ability of each scoring method to distinguish between sets of samples, we utilize three metrics: the Kolmogorov–Smirnov (KS) test, the area under the receiver operating characteristic curve (AUROC), and the false positive rate (FPR) at a 95\% true positive rate (TPR). Each metric provides unique insights into method performance in distinguishing between sample types.

The two-sample KS test is traditionally used to assess whether two data samples come from the same distribution by posing the null hypothesis that both samples come from populations with the same distribution. It may also be used to test the extent to which two underlying probability distributions differ by measuring the maximum distance between the cumulative distribution functions (CDFs) of two score distributions. This distance is one way to quantify the separability of score distributions associated with distinct sample types, where higher KS values indicate greater dissimilarity between distributions, thus signifying better performance in distinguishing these sample types.

We also use the AUROC to quantify the separability of score distributions. A receiver operating characteristic (ROC) curve is a graph that shows the performance of a model by plotting the TPR against the FPR at different threshold values. The AUROC is a widely used metric that evaluates the performance of a binary classifier across different decision thresholds. Here, we use the AUROC to measure the degree of overlap between score distributions for different sample types. An AUROC of 1.0 signifies perfect separability, while an AUROC of 0.5 suggests no discriminative power between the classes.

Finally, the FPR at a fixed TPR of 95\% assesses the rate at which a model incorrectly labels a sample as a positive point (i.e., OOD, low-accuracy, or unfamiliar) when it is, in fact, a negative point (i.e., in-distribution, high-accuracy, or familiar). By fixing the TPR at 95\%, we ensure the competency estimator maintains high sensitivity to anomalies or misclassifications. A lower FPR at this threshold implies that the model is better at avoiding false alarms while still accurately identifying true OOD, low-accuracy, and unfamiliar samples. This metric is valuable because false positives can reduce trustworthiness.


\subsection{Overall Competency Scores}
\label{subsec:app-analysis-overall}

\begin{figure*}[h!]
  \centering
  \includegraphics[width=0.9\linewidth]{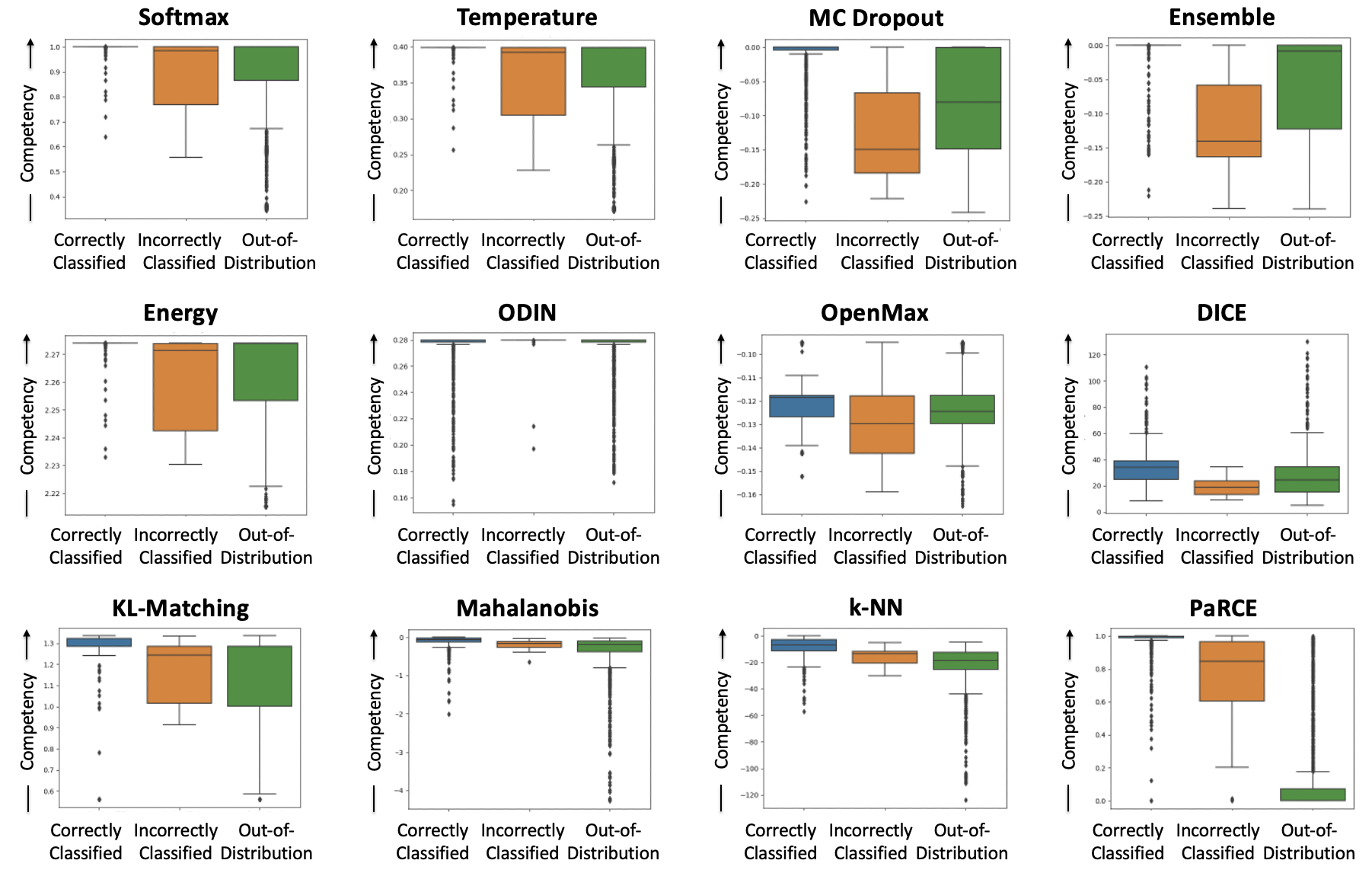}
  \caption{Box plots capturing the score distributions for correctly classified, incorrectly classified, and OOD samples in the \textit{lunar} dataset, where scores are generated by the Maximum Softmax Probability (MSP) baseline, the calibrated MSP with Temperature Scaling \cite{guo}, Monte Carlo (MC) Dropout \cite{dropout}, Ensembling \cite{lakshminarayanan_simple_2017}, the Energy Score \cite{liu_energy-based_2020}, ODIN \cite{liang_enhancing_2020}, OpenMax \cite{openmax}, DICE \cite{dice}, KL-Matching \cite{kl_matching}, the Mahalanobis Distance \cite{lee-2018}, k-Nearest Neighbors (k-NN) \cite{sun_out--distribution_2022}, and our PaRCE method. We expect correctly classified images to be assigned high competency scores, while misclassified and OOD samples are expected to have lower associated scores.}
  \label{fig:box-eval1-lunar}
\end{figure*}

\begin{figure*}[h!]
  \centering
  \includegraphics[width=0.9\linewidth]{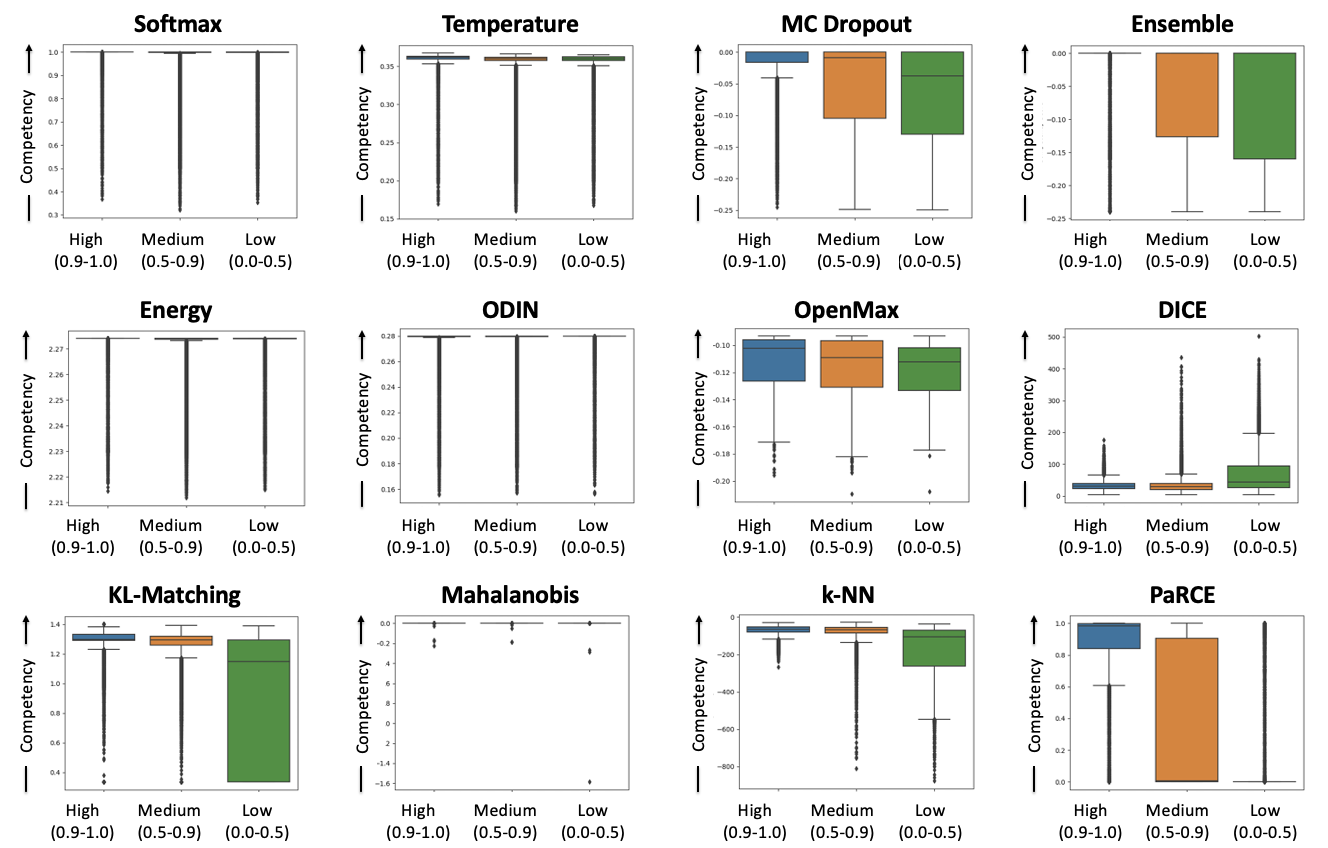}
  \caption{Box plots capturing the score distributions for images with high, medium, and low accuracy visual modifications in the \textit{lunar} dataset, where scores are generated by the Maximum Softmax Probability (MSP) baseline, the calibrated MSP with Temperature Scaling \cite{guo}, Monte Carlo (MC) Dropout \cite{dropout}, Ensembling \cite{lakshminarayanan_simple_2017}, the Energy Score \cite{liu_energy-based_2020}, ODIN \cite{liang_enhancing_2020}, OpenMax \cite{openmax}, DICE \cite{dice}, KL-Matching \cite{kl_matching}, the Mahalanobis Distance \cite{lee-2018}, k-Nearest Neighbors (k-NN) \cite{sun_out--distribution_2022}, and our PaRCE method. We expect high accuracy images to be assigned high competency scores, while medium and low accuracy samples are expected to have lower associated scores.}
  \label{fig:box-eval2-lunar}
\end{figure*}

\begin{figure*}[h!]
  \centering
  \includegraphics[width=0.9\linewidth]{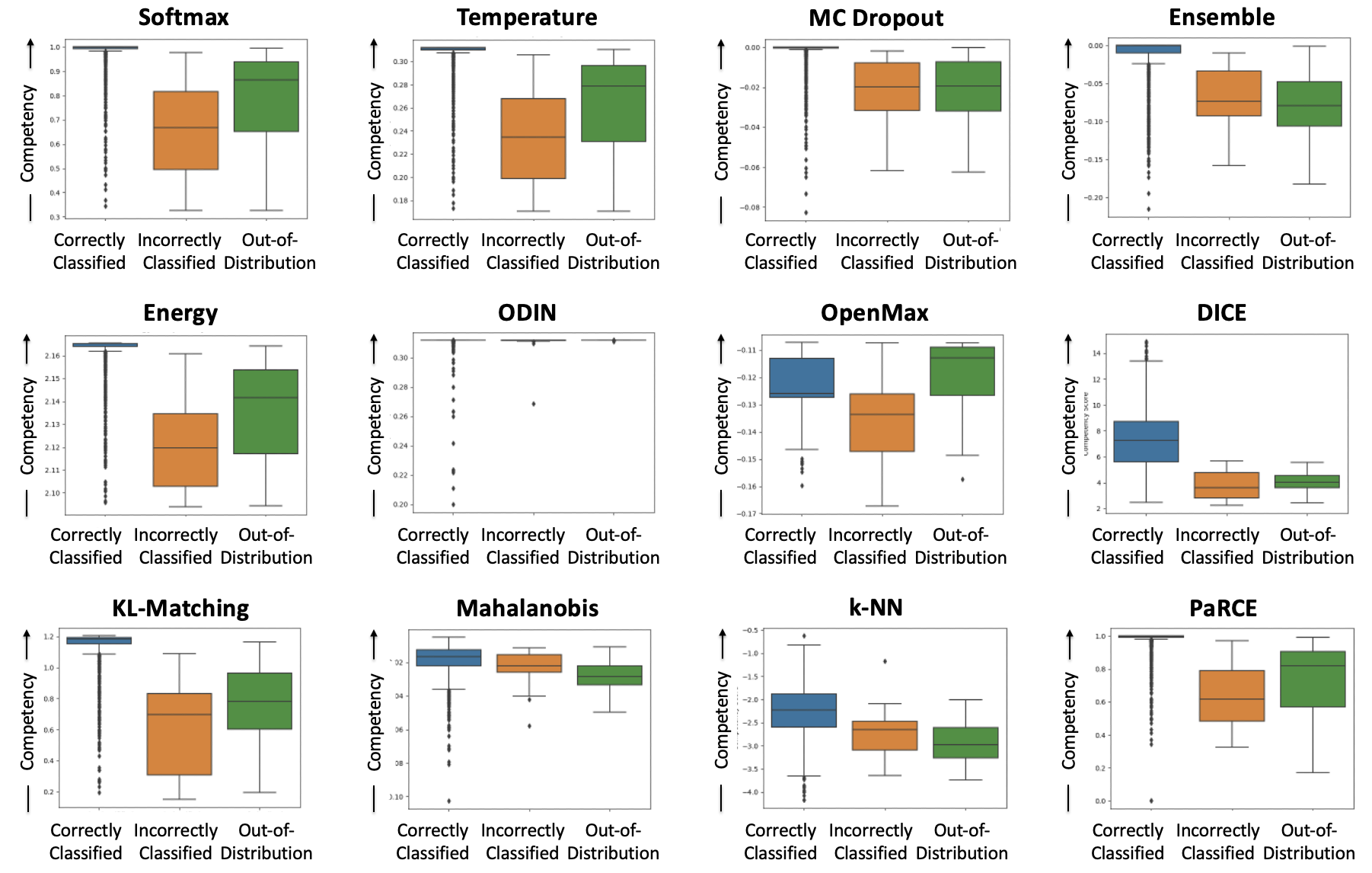}
  \caption{Box plots capturing the score distributions for correctly classified, incorrectly classified, and OOD samples in the \textit{speed limit signs} dataset, where scores are generated by the Maximum Softmax Probability (MSP) baseline, the calibrated MSP with Temperature Scaling \cite{guo}, Monte Carlo (MC) Dropout \cite{dropout}, Ensembling \cite{lakshminarayanan_simple_2017}, the Energy Score \cite{liu_energy-based_2020}, ODIN \cite{liang_enhancing_2020}, OpenMax \cite{openmax}, DICE \cite{dice}, KL-Matching \cite{kl_matching}, the Mahalanobis Distance \cite{lee-2018}, k-Nearest Neighbors (k-NN) \cite{sun_out--distribution_2022}, and our PaRCE method. We expect correctly classified images to be assigned high competency scores, while misclassified and OOD samples are expected to have lower associated scores.}
  \label{fig:box-eval1-speed}
\end{figure*}

\begin{figure*}[h!]
  \centering
  \includegraphics[width=0.9\linewidth]{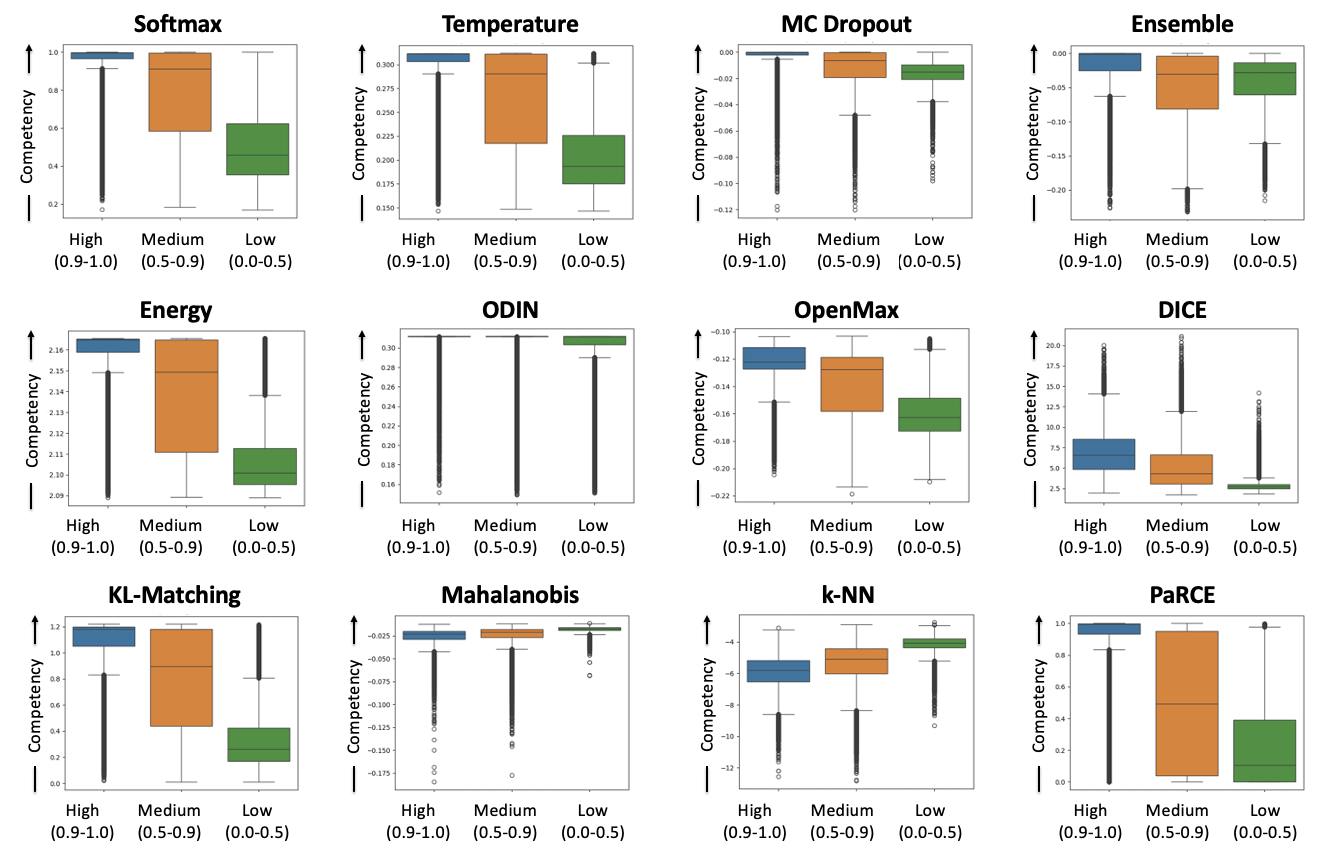}
  \caption{Box plots capturing the score distributions for images with high, medium, and low accuracy modifications in the \textit{speed limit signs} dataset, where scores are generated by the Maximum Softmax Probability (MSP) baseline, the calibrated MSP with Temperature Scaling \cite{guo}, Monte Carlo (MC) Dropout \cite{dropout}, Ensembling \cite{lakshminarayanan_simple_2017}, the Energy Score \cite{liu_energy-based_2020}, ODIN \cite{liang_enhancing_2020}, OpenMax \cite{openmax}, DICE \cite{dice}, KL-Matching \cite{kl_matching}, the Mahalanobis Distance \cite{lee-2018}, k-Nearest Neighbors (k-NN) \cite{sun_out--distribution_2022}, and our PaRCE method. We expect high accuracy images to be assigned high competency scores, while medium and low accuracy samples are expected to have lower associated scores.}
  \label{fig:box-eval2-speed}
\end{figure*}

\begin{figure*}[h!]
  \centering
  \includegraphics[width=0.9\linewidth]{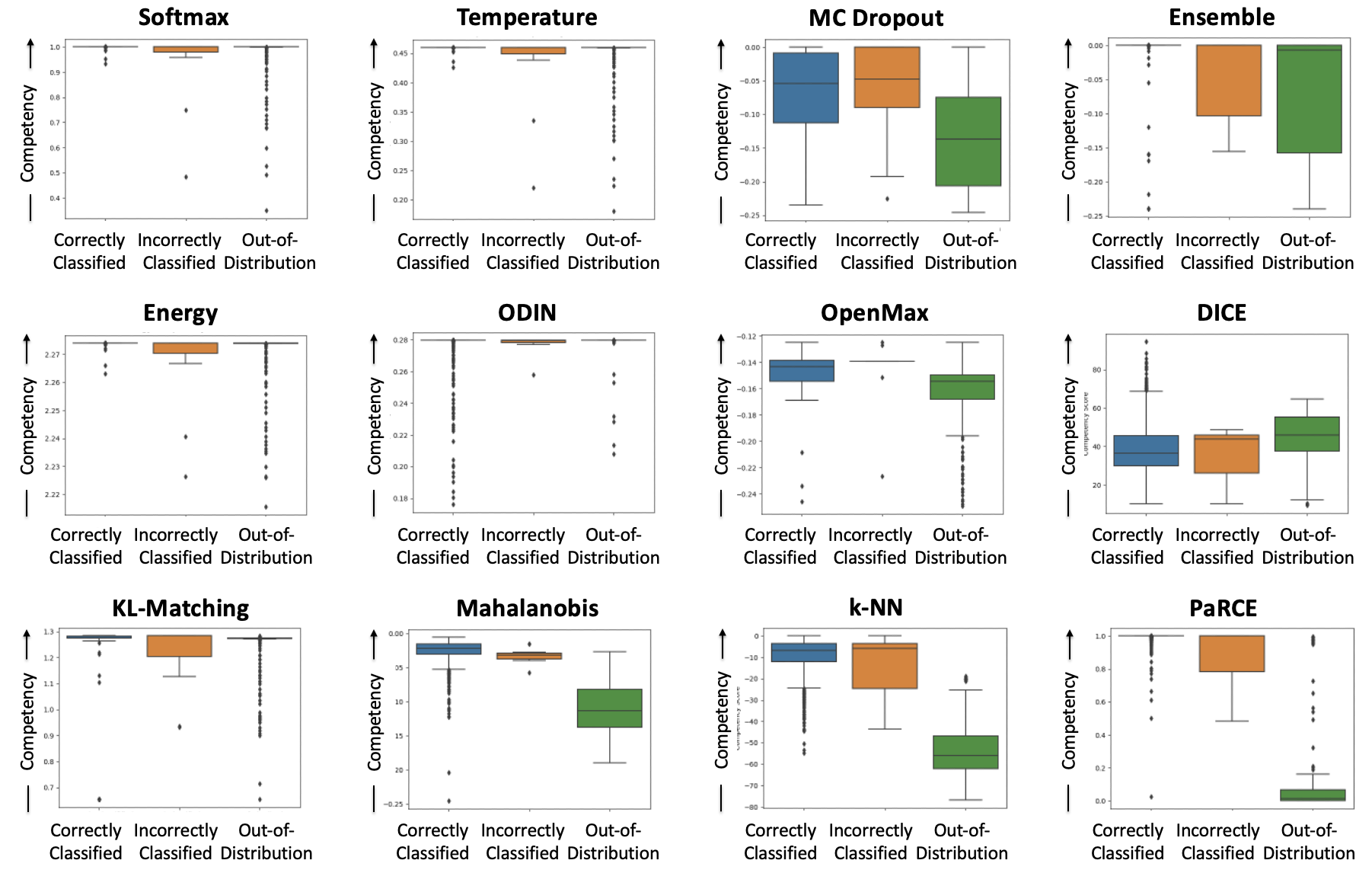}
  \caption{Box plots capturing the score distributions for correctly classified, incorrectly classified, and OOD samples in the \textit{outdoor park} dataset, where scores are generated by the Maximum Softmax Probability (MSP) baseline, the calibrated MSP with Temperature Scaling \cite{guo}, Monte Carlo (MC) Dropout \cite{dropout}, Ensembling \cite{lakshminarayanan_simple_2017}, the Energy Score \cite{liu_energy-based_2020}, ODIN \cite{liang_enhancing_2020}, OpenMax \cite{openmax}, DICE \cite{dice}, KL-Matching \cite{kl_matching}, the Mahalanobis Distance \cite{lee-2018}, k-Nearest Neighbors (k-NN) \cite{sun_out--distribution_2022}, and our PaRCE method. We expect correctly classified images to be assigned high competency scores, while misclassified and OOD samples are expected to have lower associated scores.}
  \label{fig:box-eval1-park}
\end{figure*}

\begin{figure*}[h!]
  \centering
  \includegraphics[width=0.9\linewidth]{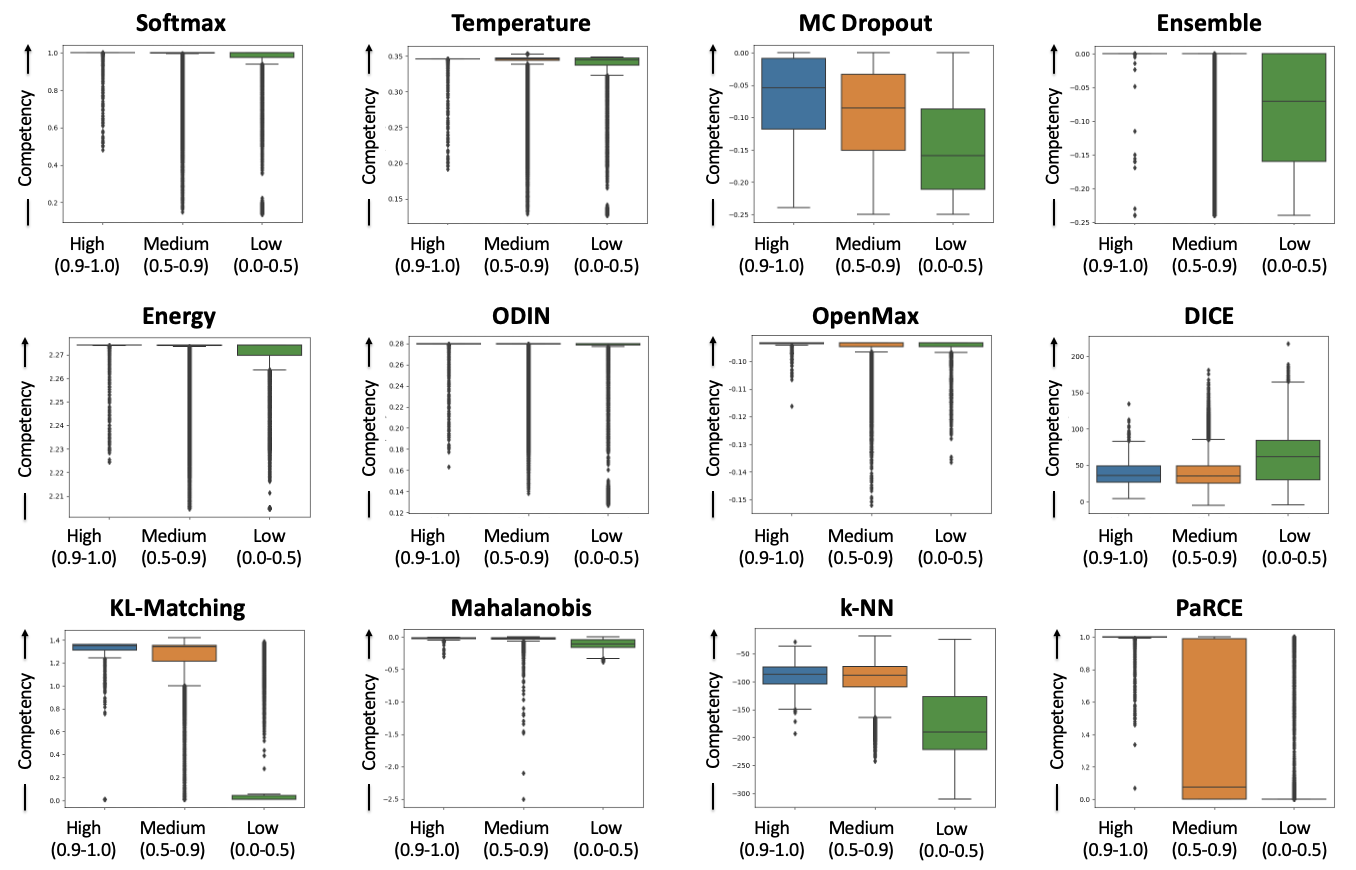}
  \caption{Box plots capturing the score distributions for images with high, medium, and low accuracy visual modifications in the \textit{outdoor park} dataset, where scores are generated by the Maximum Softmax Probability (MSP) baseline, the calibrated MSP with Temperature Scaling \cite{guo}, Monte Carlo (MC) Dropout \cite{dropout}, Ensembling \cite{lakshminarayanan_simple_2017}, the Energy Score \cite{liu_energy-based_2020}, ODIN \cite{liang_enhancing_2020}, OpenMax \cite{openmax}, DICE \cite{dice}, KL-Matching \cite{kl_matching}, the Mahalanobis Distance \cite{lee-2018}, k-Nearest Neighbors (k-NN) \cite{sun_out--distribution_2022}, and our PaRCE method. We expect high accuracy images to be assigned high competency scores, while medium and low accuracy samples are expected to have lower associated scores.}
  \label{fig:box-eval2-park}
\end{figure*}

In this section, we provide the score distributions for correctly classified, misclassified, and OOD samples across various methods for assessing overall model competency. \cref{fig:box-eval1-lunar} displays box plots of the distributions for the lunar dataset, \cref{fig:box-eval1-speed} displays plots for the speed limit signs dataset, and \cref{fig:box-eval1-park} displays plots for the outdoor park dataset. We also provide the score distributions for samples with high, medium, and low accuracy modifications across the same methods for assessing overall model competency. \cref{fig:box-eval2-lunar} displays box plots of the distributions for the lunar dataset, \cref{fig:box-eval2-speed} displays plots for the speed limit signs dataset, and \cref{fig:box-eval2-park} displays plots for the outdoor park dataset.

\subsection{Regional Competency Images}
\label{subsec:app-analysis-regional}

\begin{figure*}[h!]
  \centering
  \includegraphics[width=0.9\linewidth]{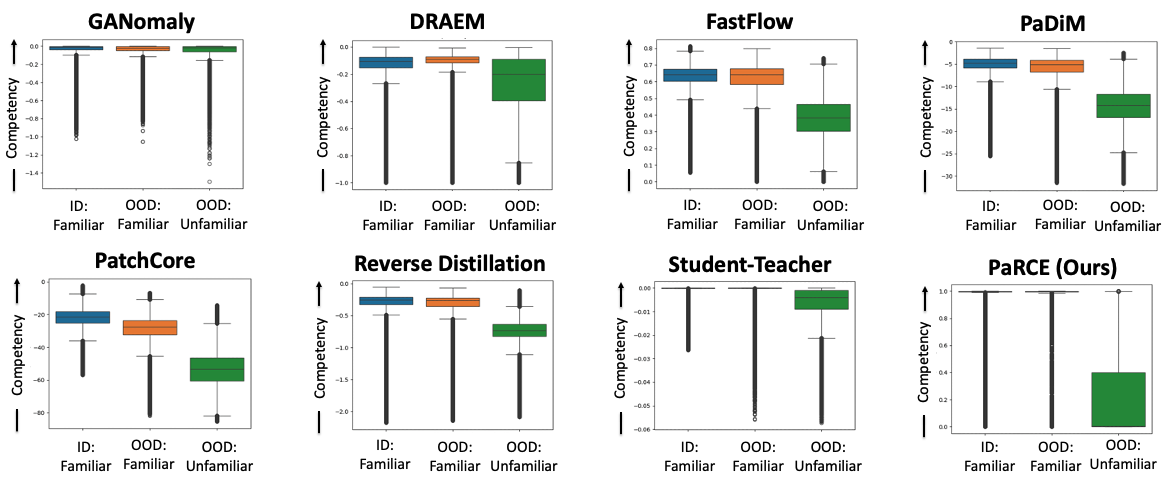}
  \caption{Box plots capturing the score distributions for all pixels of in-distribution (ID) images, familiar pixels of OOD images, and unfamiliar pixels of OOD images in the \textit{lunar} dataset, where scores are generated by GANomaly \cite{ganomaly}, DRAEM \cite{zavrtanik_draem_2021}, FastFlow \cite{fastflow}, PaDiM \cite{del_bimbo_padim_2021}, PatchCore \cite{patchcore}, Reverse Distillation \cite{reverse}, Student-Teacher Feature Pyramid Matching \cite{stfpm}, and our PaRCE method. We expect high competency scores for familiar pixels (from both ID and OOD images) and lower scores for unfamiliar pixels.}
  \label{fig:box-eval3-lunar}
\end{figure*}

\begin{figure*}[h!]
  \centering
  \includegraphics[width=0.9\linewidth]{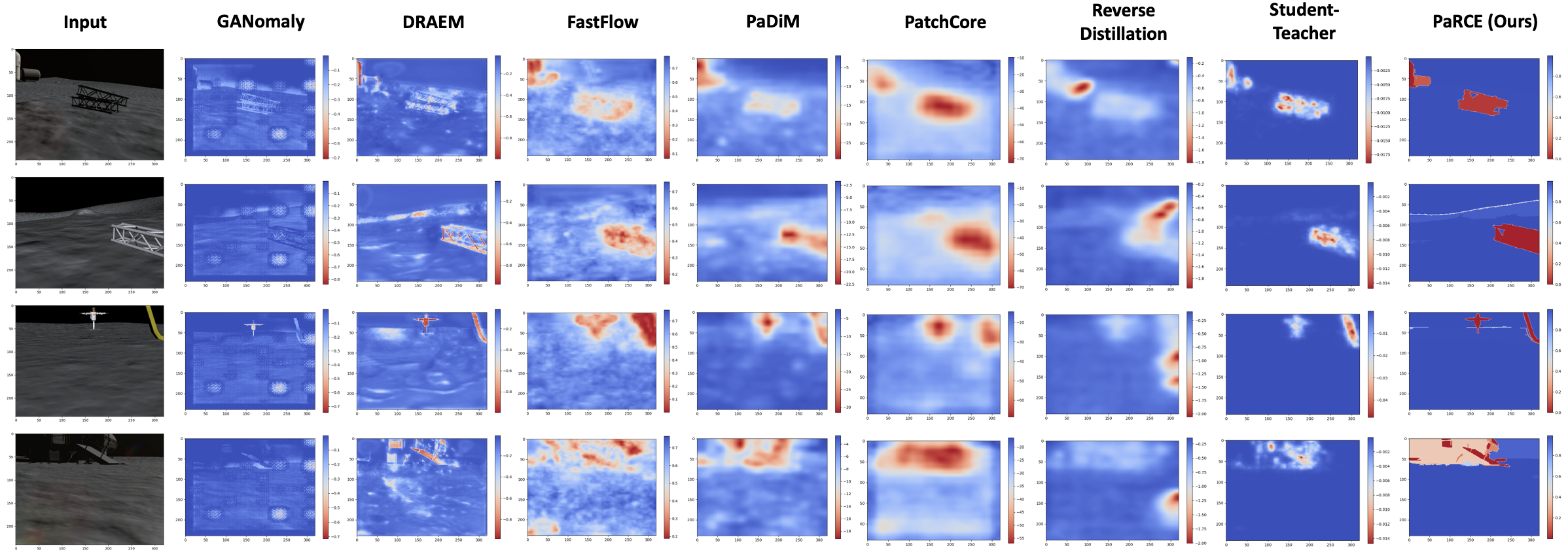}
  \caption{A comparison of the regional competency images obtained for example OOD images in the \textit{lunar} dataset, for which the sky and lunar surface are familiar to the perception model while astronauts and human-made structures are unfamiliar. A pixel assigned a high competency score appears more blue, while a pixel assigned a low score appears more red. We expect more red regions to correspond to unfamiliar objects (i.e., an astronaut or human-made structure) while more blue regions should correspond to familiar structures.}
  \label{fig:regional-lunar}
\end{figure*}

\begin{figure*}[h!]
  \centering
  \includegraphics[width=0.9\linewidth]{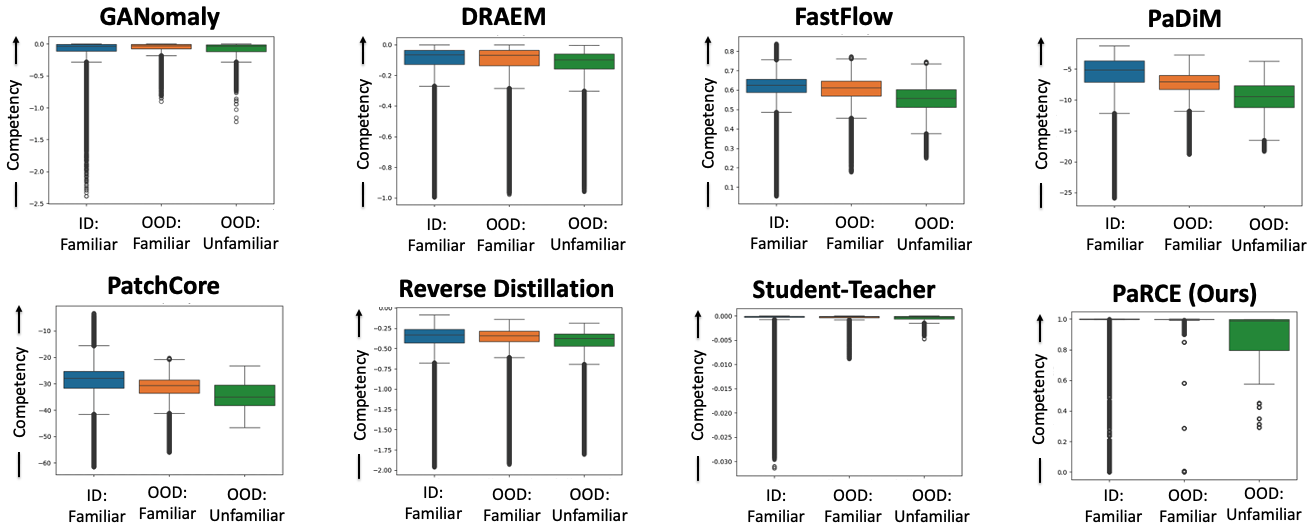}
  \caption{Box plots capturing the score distributions for all pixels of in-distribution (ID) images, familiar pixels of OOD images, and unfamiliar pixels of OOD images in the \textit{speed limit signs} dataset, where scores are generated by GANomaly \cite{ganomaly}, DRAEM \cite{zavrtanik_draem_2021}, FastFlow \cite{fastflow}, PaDiM \cite{del_bimbo_padim_2021}, PatchCore \cite{patchcore}, Reverse Distillation \cite{reverse}, Student-Teacher Feature Pyramid Matching \cite{stfpm}, and our PaRCE method. We expect high competency scores for familiar pixels (from both ID and OOD images) and lower scores for unfamiliar pixels.}
  \label{fig:box-eval3-speed}
\end{figure*}

\begin{figure*}[h!]
  \centering
  \includegraphics[width=0.9\linewidth]{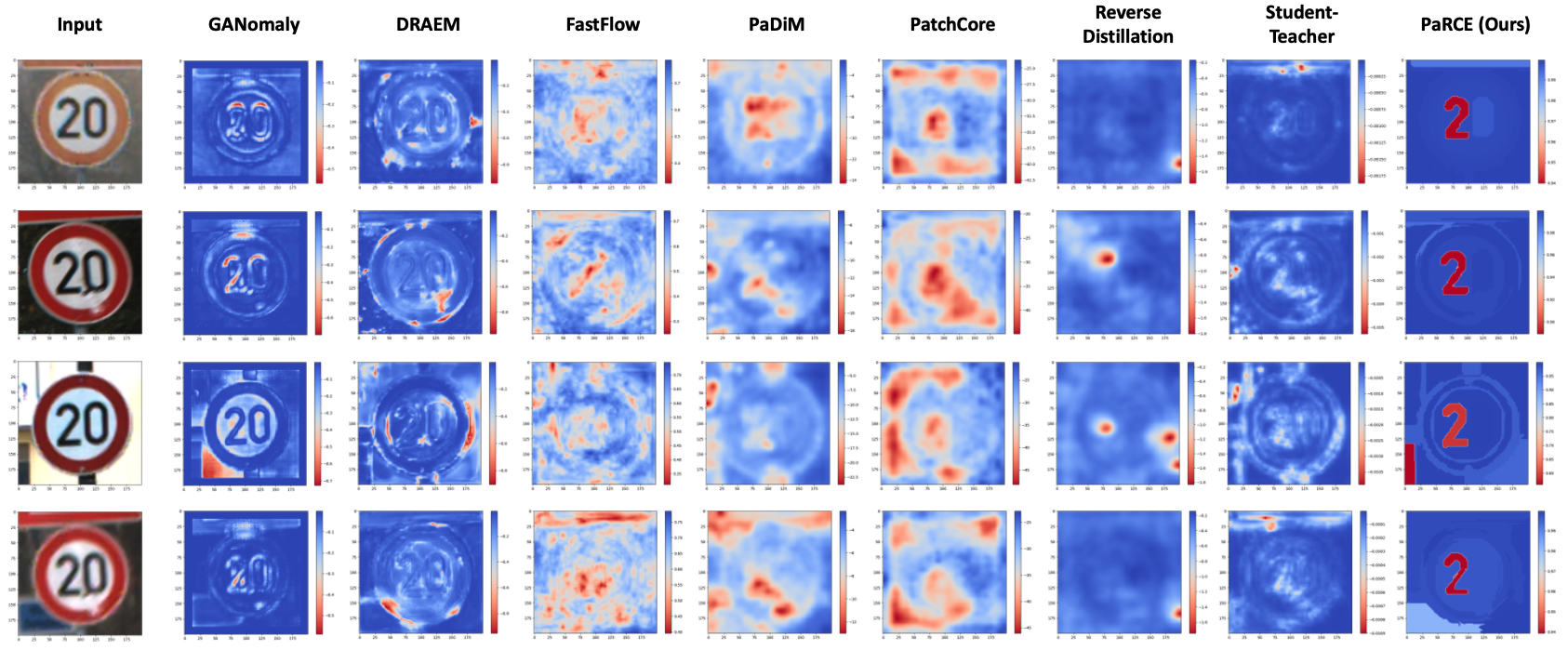}
  \caption{A comparison of the regional competency images obtained for example OOD images in the \textit{speed limit signs} dataset, for which speeds 30-120 km/hr are familiar to the perception model while the speed 20 km/hr is unfamiliar. A pixel assigned a high competency score appears more blue, while a pixel assigned a low score appears more red. We expect more red regions to correspond to regions associated with the unseen class (i.e., the number 2) while more blue regions should correspond to other parts of the traffic sign.}
  \label{fig:regional-speed}
\end{figure*}

\begin{figure*}[h!]
  \centering
  \includegraphics[width=0.9\linewidth]{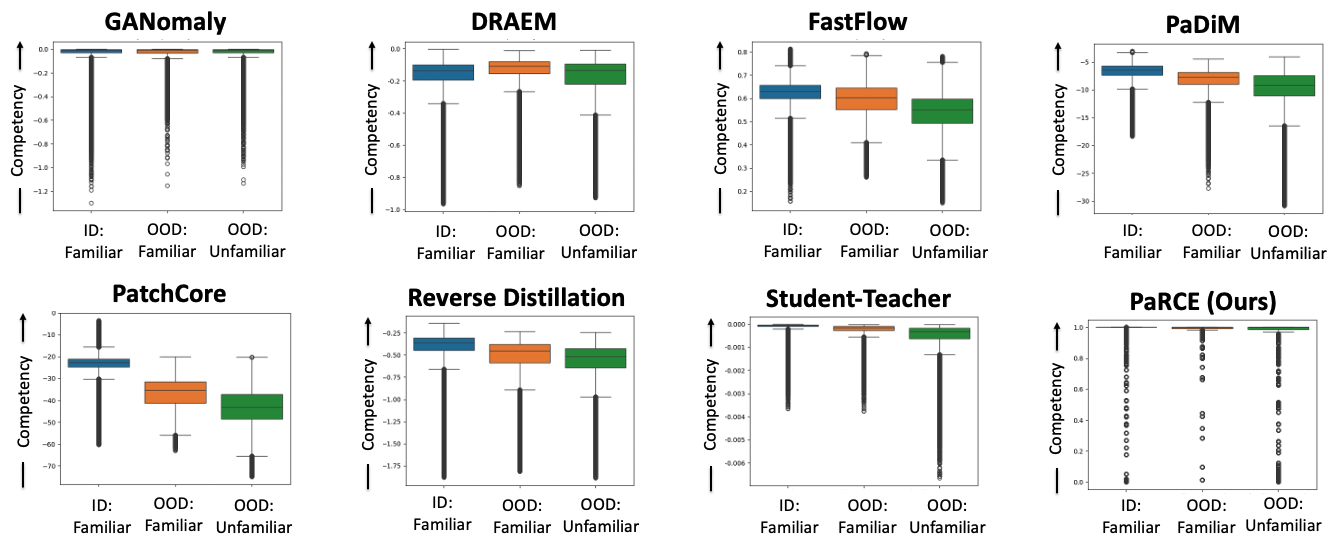}
  \caption{Box plots capturing the score distributions for all pixels of in-distribution (ID) images, familiar pixels of OOD images, and unfamiliar pixels of OOD images in the \textit{outdoor park} dataset, where scores are generated by GANomaly \cite{ganomaly}, DRAEM \cite{zavrtanik_draem_2021}, FastFlow \cite{fastflow}, PaDiM \cite{del_bimbo_padim_2021}, PatchCore \cite{patchcore}, Reverse Distillation \cite{reverse}, Student-Teacher Feature Pyramid Matching \cite{stfpm}, and our PaRCE method. We expect high competency scores for familiar pixels (from both ID and OOD images) and lower scores for unfamiliar pixels.}
  \label{fig:box-eval3-park}
\end{figure*}

\begin{figure*}[h!]
  \centering
  \includegraphics[width=0.9\linewidth]{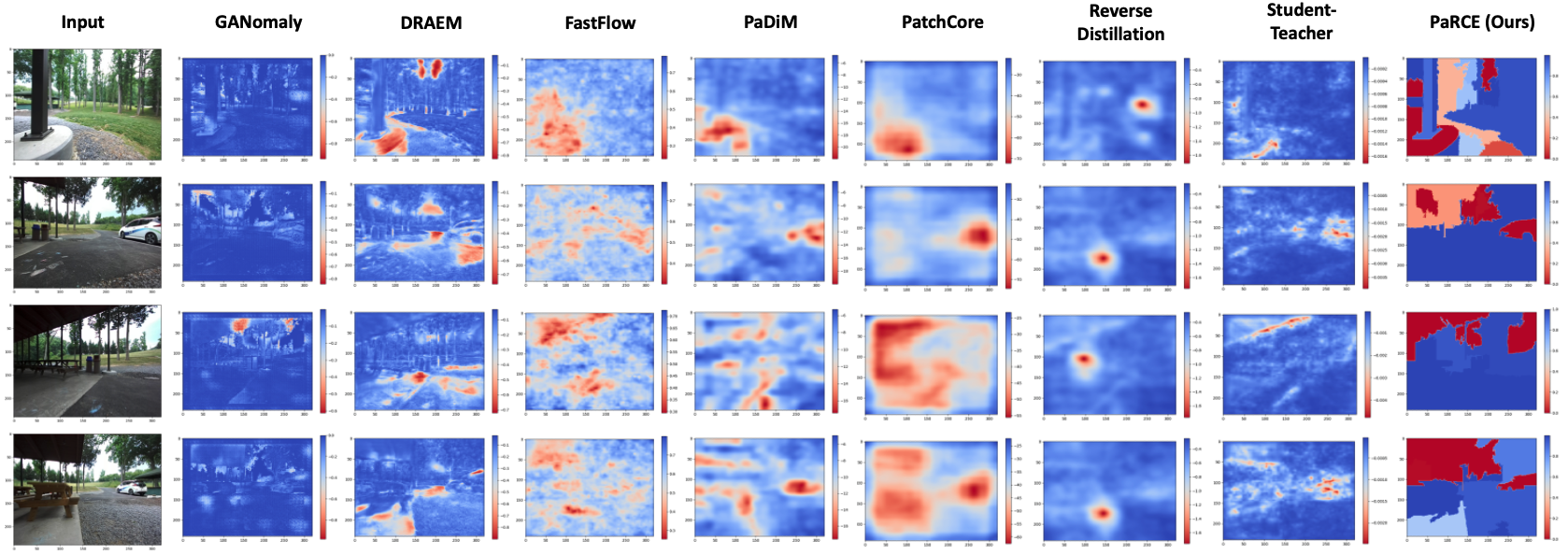}
  \caption{A comparison of the regional competency images obtained for example OOD images in the \textit{outdoor park} dataset, for which the grassy and forested regions are familiar to the perception model while regions around the pavilion are unfamiliar. A pixel assigned a high competency score appears more blue, while a pixel assigned a low score appears more red. We expect more red regions to correspond to unexplored regions in the environment (i.e., areas around the pavilion) while more blue regions should correspond to familiar regions.}
  \label{fig:regional-park}
\end{figure*}

In this section, we provide the score distributions for familiar regions in in-distribution (ID) images, familiar regions in OOD images, and unfamiliar regions in OOD images across various methods for assessing regional model competency. \cref{fig:box-eval3-lunar} displays box plots of the distributions for the lunar dataset, \cref{fig:box-eval3-speed} displays plots for the speed limit signs dataset, and \cref{fig:box-eval3-park} displays plots for the outdoor park dataset. We also provide examples of the competency images generated by each of the same methods for assessing regional model competency. \cref{fig:regional-lunar} displays example images for the lunar dataset, \cref{fig:regional-speed} displays examples for the speed limit signs dataset, and \cref{fig:regional-park} displays examples for the outdoor park dataset.

\newpage

\end{document}